\documentclass[times, review, 10pt]{elsarticle}

\usepackage[normalem]{ulem}
\usepackage{graphicx}
\usepackage{amssymb}
\usepackage{dsfont}
\usepackage{caption}
\usepackage{subcaption}
\usepackage{amsmath} 
\usepackage{adjustbox}
\usepackage{hyperref}
\usepackage{tabularx}
\usepackage{multirow}
\usepackage{float}
\usepackage{xcolor}
\usepackage{placeins}

\useunder{\uline}{\ul}{}
\newcolumntype{U}{>{\centering\arraybackslash}p{0.1\textwidth}}
\newcolumntype{V}{>{\centering\arraybackslash}p{0.3\textwidth}}
\newcolumntype{W}{>{\centering\arraybackslash}p{0.03\textwidth}}
\newcolumntype{Y}{>{\centering\arraybackslash}p{0.2\textwidth}}
\newcolumntype{Z}{>{\centering\arraybackslash}X} 

\begin{document}
\begin{frontmatter}

\title{Fine-Tuning Video-Text Contrastive Model for Primate Behavior Retrieval from Unlabeled Raw Videos}

\author{
  Giulio Cesare Mastrocinque Santo$^1$, Patrícia Izar$^2$, Irene Delval$^2$, Victor de Napole Gregolin$^3$, Nina S. T. Hirata$^1$  \\
  $^1$ \scriptsize  Institute of Mathematics and Statistics, University of São Paulo (IME-USP), Rua do Matão, 1010, São Paulo, 05508-090, São Paulo, Brazil \\
  $^2$ \scriptsize  Department of Experimental Psychology, Institute of Psychology, University of São Paulo (IP-USP), Av. Professor Mello Moraes, 1721, São Paulo, 05508-030, São Paulo, Brazil \\
  $^3$ \scriptsize  Institute of Biosciences, University of São Paulo (IB-USP), Rua do Matão, 321, São Paulo, 05508-090, São Paulo, Brazil
}

%% Abstract
\begin{abstract}
Video recordings of nonhuman primates in their natural habitat are a common source for studying their behavior in the wild. We fine-tune pre-trained video-text foundational models for the specific domain of capuchin monkeys, with the goal of developing useful computational models to help researchers to retrieve useful clips from videos. We focus on the challenging problem of training a model based solely on raw, unlabeled video footage, using weak audio descriptions sometimes provided by field collaborators. We leverage recent advances in Multimodal Large Language Models (MLLMs) and Vision-Language Models (VLMs) to address the extremely noisy nature of both video and audio content. Specifically, we propose a two-folded approach: an agentic data treatment pipeline and a fine-tuning process. The data processing pipeline automatically extracts clean and semantically aligned video-text pairs from the raw videos, which are subsequently used to fine-tune a pre-trained Microsoft's X-CLIP model through Low-Rank Adaptation (LoRA). We obtained an uplift in $Hits@5$ of $167\%$ for the 16 frames model and an uplift of $114\%$ for the 8 frame model on our domain data. Moreover, based on $NDCG@K$ results, our model is able to rank well most of the considered behaviors, while the tested raw pre-trained models are not able to rank them at all. The code will be made available upon acceptance.
\end{abstract}

%% Keywords
\begin{keyword} 
video-text contrastive models \sep agentic data treatment  \sep Low-Rank Adaptation \sep video-text retrieval \sep zero-shot classification \sep video in the wild \sep capuchin monkey behavior
\end{keyword}

\end{frontmatter}

%% main text
\section{Introduction}
\label{sec:introduction}
Primate behavior has long been a central focus of research in disciplines like comparative psychology, anthropology, and biology. At the Laboratory of Ethology, Social Interactions, and Development (LEDIS) of the Institute of Psychology of the University of S\~{a}o Paulo (IP-USP), psychologists and biologists investigate the roots of human behavior by examining the developmental trajectories of capuchin monkeys, observing individuals from birth through adulthood. 

These studies aim to uncover, for example, how capuchins acquire tool-use behaviors -- such as using rocks to crack open nuts -- and how such technical traditions emerge \cite{ref_monkeys_tradition}, \cite{ref_monkeys_objects_tools}; how they develop object manipulation skills over time \cite{ref_early_object_manipulation}; and when and how personality traits begin to form \cite{ref_personality_assessment}. Another example of research focus is the study about the mechanisms and evolutionary origins of behavioral responses to death  \cite{ref_carrying_the_dead}.

In 2013, a longitudinal study of a wild group of capuchin monkeys (\textit{Sapajus xanthosternos}) begun at the Una Biological Reserve in Bahia, Brazil ($15^{\circ}6'-12'\,\text{S}$ and $39^{\circ}02'-12'\,\text{W}$). Since then, videos of all individuals from birth to three years of age have been collected weekly through focal follows \cite{ref_sampling_focal_follows}. As part of a larger monitoring effort by the LEDIS group, this footage aimed to document behaviors relevant to specific research questions such as the above mentioned. For some individuals, the filming continued until their fifth year of life, providing a rich dataset on early behavioral development.

Such behaviors, however, are often unpredictable and rarely observed, and this is just part of the challenge. Most recordings are noisy for several reasons. First, due to the long duration of the project, changes in camera equipment have resulted in inconsistent video quality. Second, in many cases, monkeys do not appear at all, appear only briefly, or are hidden by dense vegetation. Finally, unfavorable weather or excessive camera movement can also degrade video quality. As a result, only a fraction of the videos actually contain relevant behaviors of interest, and the researchers are therefore left with a massive volume of noisy footage that must be manually reviewed to extract data suitable for meaningful analysis.

It is natural, therefore, to think about the use of computational techniques to address such a challenging problem, and this is the central goal of our work. With major advances in deep learning, particularly in the areas of Computer Vision (CV) and Natural Language Processing (NLP), there are multiple possible paths that we could explore. As previously mentioned, researchers in the LEDIS group have been analyzing these videos for years, producing a modest but valuable set of annotated data that we could use to train specialized models. Another option would be to train models, such as classification or detection models, with the specific purpose of filtering videos, retaining only those featuring actual monkeys -- or even specific individuals -- to reduce noise and facilitate analysis. 

However, two aspects of the data caught our attention. First, when interesting behaviors are observed, it can occur that a field collaborator verbally describes what is happening, assigning a spoken description to the video. Those verbal descriptions are not very common and usually are inaccurate and non-technical, but a minority of them can serve as valuable video captions. Second, the overall volume of available videos is huge. Although in this paper we accessed only a subsample of the dataset, the existence of a larger dataset enables training of more complex models in future work. 

Those two aspects led us to adopt a more ambitious approach: training a foundational model directly from raw video data, without relying on manual annotations. Relying on manual annotation creates a dependency on a labor-intensive and exhausting process, which is difficult to sustain over time and limits the ability to update models with newer data. Moreover, we strongly believe that building a model from raw data is now feasible due to the growing availability of powerful pre-trained models and due to the rapid progress in multimodal learning, as proven by models like CLIP \cite{ref_clip} and emerging Multimodal Large Language Models (MLLMs) such as LLaMA \cite{ref_LLaMA}. 

Finally, this approach allows us to use newly available and collected data to continuously improve such a foundational model, which can be used to support several different applications, such as identifying specific behaviors from a reference video clip; retrieving behaviors from textual descriptions, filtering only videos where the monkeys actually appear or videos that contain specific actions; and serving as a backbone and a domain-specific feature extractor for training more specialized models.

To advance towards our goal, we propose a method that employs multimodal video-text models, taking advantage of the fact that the videos' audio contains some hints about the individuals being
filmed and about the type of action they are involved in. The big challenge is, of course, on how to deal with the noisy nature of the videos, and select only the representative clip-text pairs to be used to fine-tune existing models. For that purpose, we first develop an agentic data treatment pipeline based on MLLMs to produce informative training data; then, we fine-tune a pre-trained video-text model on the produced data to adapt it to our domain. We employ Low-Rank Adaptation (LoRA) \cite{ref_LoRA}, which is a Parameter-Efficient Fine-Tuning
(PEFT) method popularized for LLMs, but little explored in vision models.

Our contributions are the following.
\begin{enumerate}
    \item A novel method to automatically select semantically aligned clip-text pairs from the raw videos, without explicit supervision. We use an ethogram, which is a catalog of behaviors or animal actions commonly used in ethology, to help us filter only relevant transcripts.

    \item Successful LoRA-based fine-tuning of X-CLIP~\cite{ref_xclip} using a limited amount of domain-specific training data, demonstrated by substantial performance improvements on retrieval and zero-shot
    classification tasks when compared to several versions os raw
    X-CLIP pretrained models.

    \item Demonstration that combining (1) and (2) above improves processing of noisy videos, such as the recordings of capuchin monkeys in their natural habitat analyzed in this paper. To the best of our knowledge, this is the first work that explores behavior detection of capuchin monkeys through video-text contrastive models.
\end{enumerate}

The remainder of this paper is organized as follows. In Section \ref{sec:background}, we present the theoretical foundations adopted in the paper, as well as related works that we used as inspiration. In Section \ref{sec:method} we provide details of our method, explaining both the data treatment pipeline and the adopted fine-tuning process. Section \ref{sec:results} discusses the setup of the experiments and the results obtained, while Section \ref{sec:conclusions_and_future_work} draws the final conclusions and discusses potential future work.

\section{Background and Related Works}
\label{sec:background}

In this section, we explore some of the main concepts we apply alongside the paper. More specifically, we go through vision-text contrastive models, which is the type of model we are adopting. We also briefly review some fine-tuning techniques for vision models and give some insights into Multimodal Large Language Models (MLLMs) and LLM-based agents, which are the backbone we use to create our own data processing pipeline.

\subsection{Video-Text Contrastive Models} 
\label{subsec:video_text_models}
Video-text contrastive models are a particular type of Vision-Language Models (VLMs) that encode paired video-text inputs into a shared embedding space, such that semantically aligned (matching) pairs are mapped to nearby vectors, while unaligned pairs are mapped farther apart in the embedding vector space. Video-text models are a natural extension of image-text contrastive models, that were popularized with the emergence of models such as CLIP \cite{ref_clip}, CoCa \cite{ref_coca} and SigLip \cite{ref_siglip_1}, \cite{ref_siglip_2}. Typically, these models are dual-encoders, which means they have separate encoders for visual and textual data that produce vector representations (embeddings) of their respective modalities. Moreover, being trained with a vast amount of data, they are capable of generalizing to unseen domains, making them useful for zero-shot classification and retrieval tasks and serving as the backbone to train and adapt models without the need of a large amount of extra data. 

\begin{figure}[ht]
\centering
\includegraphics[scale=0.08]{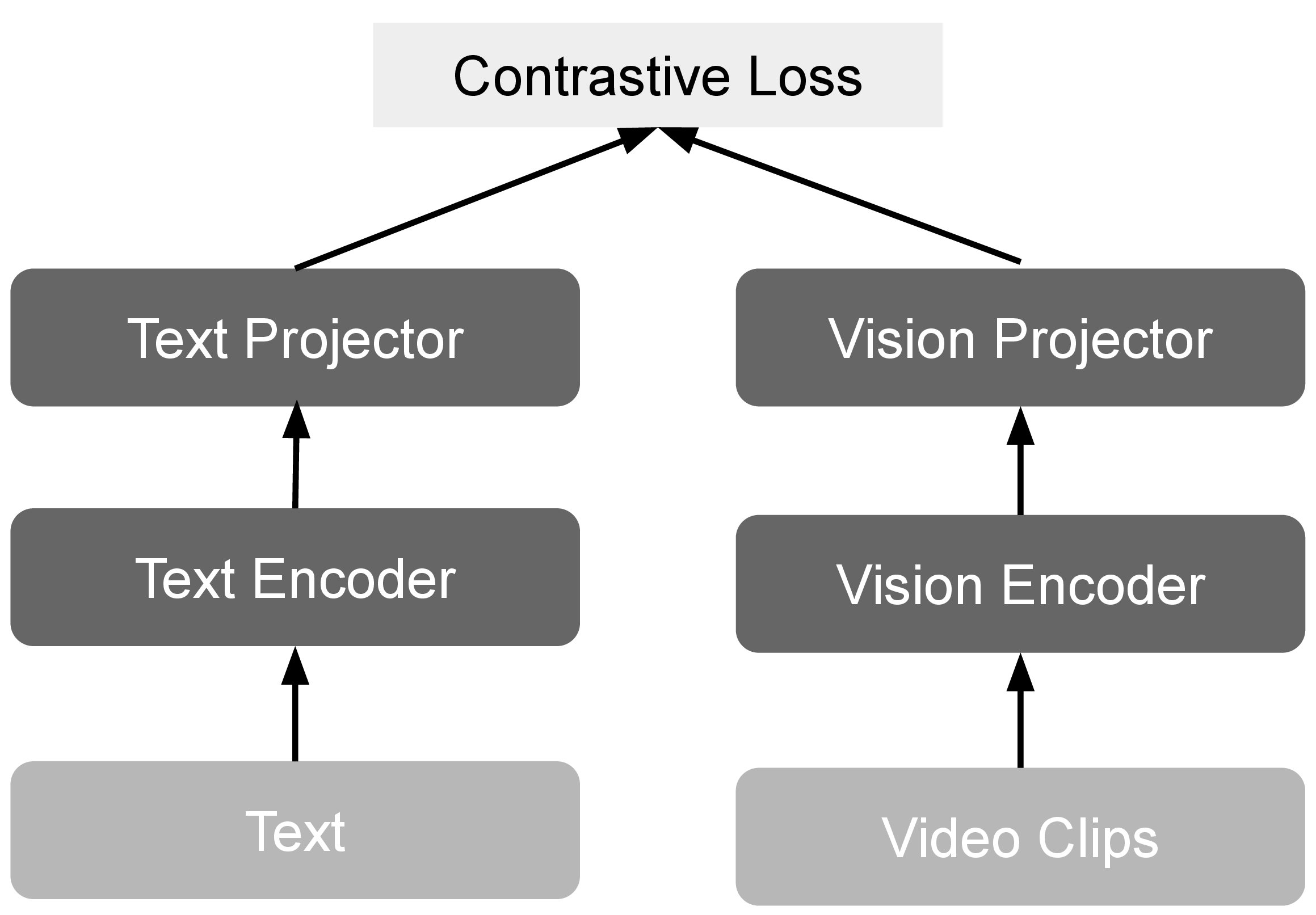}
\caption{\textbf{Video-Text Dual Encoder Architecture}.}
\label{fig:video_text_models}
\end{figure}

Video-text contrastive models also usually contain two independent Transformer based encoders, as displayed in Fig.~\ref{fig:video_text_models}. The input to these type of models is a pair $(c, t)$ of a video clip $c$ and its corresponding description $t$ (text). Both text and video inputs go, separately, into an encoder module and then into a projector module. Both projectors generate a vector (embedding) in a shared vector space, for instance, in $\mathbb{R}^d$. We denote the embedding of $t$ as $\mathbf{t}$ and the embedding of $c$ as $\mathbf{c}$.

The resulting vectors $\mathbf{c}$ and $\mathbf{t}$ are then used to calculate a contrastive loss that minimizes the cosine distance between semantically aligned video-text pairs and maximizes the distance between disjoint pairs. Because the encoders are independent, those models can naturally be used in retrieval tasks, which makes them really powerful. For example, we can pre-compute video embeddings in a large set of videos and index them in a vector database. Then, given a textual input, we can obtain the corresponding text embedding and, because text and video embeddings are in a shared vector space, we can search for the most similar video embeddings using some vector similarity metric, such as cosine similarity.

This is the case in models like CLIP \cite{ref_clip} and SigLip \cite{ref_siglip_1}. In contrast, models such as CoCa \cite{ref_coca} and SigLip-2 \cite{ref_siglip_2} optimize multiple loss functions. CoCa (Contrastive Captioner), for instance, combines contrastive loss with a captioning objective that involves next-token prediction (text generation), which adopts a cross-attention mechanism.

\subsubsection{Contrastive Loss} 
\label{subsubsec:contrastivel_loss}

The contrastive loss shown in Figure~\ref{fig:video_text_models} is an important concept and, therefore, we present it in detail here. It is computed on training batches with $N_{B}$ clip-transcript pairs $\{(c_{1}, t_{1}), (c_{2}, t_{2}),  ..., (c_{N_{B}}, t_{N_{B}})\}$. Each transcript embedding $\mathbf{t}_i$ is contrasted to each of the clip embeddings $\mathbf{c}_j$ ($j=1,\ldots,N_B$) and, conversely, each clip embedding $\mathbf{c}_j$ is contrasted to each of the transcript embeddings $\mathbf{t}_i$ ($i=1,\ldots,N_B$). For the $i$-th transcript, we would like to have $\mathbf{t}_i \approx \mathbf{c}_i$ and $\mathbf{t}_i \not\approx \mathbf{c}_j$ for $j\neq i$. Analogously, for the $j$-th clip we would like to have $\mathbf{c}_j \approx \mathbf{t}_i$ and $\mathbf{c}_j \not\approx \mathbf{t}_i$ for $i\neq j$. The cosine similarity between $\mathbf{t}_i$ and $\mathbf{c}_j$ is given by $\langle\mathbf{t}_{i}, \mathbf{c}_{j}\rangle = \langle\mathbf{c}_{j}, \mathbf{t}_{i}\rangle = \frac{\mathbf{t}_{i} \cdot \mathbf{c}_{j}}{||\mathbf{t}_{i}|| \cdot ||\mathbf{c}_{j}||}$, $i,j=1,\ldots,N_B$.
 
Thus, using the softmax function, we can define, for $i=1\ldots,N_B$, the following prediction probabilities:
\begin{equation}
\label{eq:y_tc}
    \hat{y}_{t_i} = \frac{exp(\frac{\langle\mathbf{t}_{i}, \mathbf{c}_{i}\rangle}{\tau})}{\sum^{N_{B}}_{j=1}exp(\frac{\langle\mathbf{t}_{i}, \mathbf{c}_{j}\rangle}{\tau})}
\end{equation}
and, similarly, for $j=1\ldots,N_B$:
\begin{equation}
\label{eq:y_ct}
    \hat{y}_{c_j} = \frac{exp(\frac{\langle\mathbf{c}_{j}, \mathbf{t}_{j}\rangle}{\tau})}{\sum^{N_{B}}_{i=1}exp(\frac{\langle\mathbf{c}_{j}, \mathbf{t}_{i}\rangle}{\tau})}
\end{equation}

Notice that $\hat{y}_{t_i}\approx 1$ implies $\mathbf{t}_i \approx \mathbf{c}_i$ and, analogously, $\hat{y}_{c_j}\approx 1$ implies $\mathbf{c}_j \approx \mathbf{t}_j$. Temperature $\tau$ controls the prediction probabilities distributions. Thus, the contrastive loss over the batch of $N_B$ clip-transcript pairs can be viewed as a multi-class classification task with $N_{B}$ classes, and therefore a suitable loss function is the cross-entropy loss. Using $\mathds{1}_{\{i,j\}}=1 \Leftrightarrow i=j$, one can write the cross-entropy losses treating video clips against transcriptions and transcriptions against video clips, respectively, as 
\begin{equation}
\label{eq:ltc}
    L_{tc} = -\frac{1}{N_{B}} \sum^{N_{B}}_{i=1}\sum^{N_{B}}_{j=1}\mathds{1}_{\{i,j\}}log(\hat{y}_{t_i}) = -\frac{1}{N_{B}} \sum^{N_{B}}_{i=1} log(\hat{y}_{t_i})
\end{equation}
and
\begin{equation}
\label{eq:lct}
    L_{ct} = -\frac{1}{N_{B}} \sum^{N_{B}}_{j=1}\sum^{N_{B}}_{i=1}\mathds{1}_{\{j,i\}}log(\hat{y}_{c_j}) = -\frac{1}{N_{B}} \sum^{N_{B}}_{j=1} log(\hat{y}_{c_j})
\end{equation}

The contrastive loss is given by Equation~(\ref{eq:contrastive_loss}) as a symmetric loss that combines $L_{ct}$ and $L_{tc}$.
\begin{equation}
\label{eq:contrastive_loss}
    CL(c, t) = (L_{ct} + L_{tc})/2
\end{equation}

Notice that the denominator of the predicted probabilities $\hat{y}_{t_i}$ and $\hat{y}_{c_j}$ are different. In $\hat{y}_{t_i}$ we normalize the logit $\langle\mathbf{t}_{i}, \mathbf{c}_{j}\rangle$ considering every clip in the batch, while in $\hat{y}_{c_j}$ we normalize the logit considering every transcription in the batch instead. Consequently, if we were to use only $L_{ct}$ or $L_{tc}$ in isolation, we would bias the loss function towards the video or the transcript direction. The symmetric approach in Equation (\ref{eq:contrastive_loss}) and proposed in CLIP \cite{ref_clip} allows us to consider both clips and transcripts.

\subsubsection{Existing Models} 
\label{subsubsec:existing_models}
There are several models proposed in the literature that extend image-text models into the video domain. Some examples are CLIP4Clip \cite{ref_clip4clip}, VideoCLIP \cite{ref_videoclip}, TeachText \cite{ref_teachtext}, and X-CLIP \cite{ref_xclip}, the latter being the one adopted in this paper. CLIP4Clip is one of the first works that extends CLIP into the video domain by using it as the backbone to extract features in both text and video encoders. VideoCLIP \cite{ref_videoclip} also proposes a Transformer-based dual encoder architecture, but it addresses a natural challenge that arises in Video-Text contrastive models that is the poor alignment between the video and its corresponding description. The authors generate temporally overlapping video clips by randomly selecting a time point within the text's time interval and creating a clip of random duration around it. This approach allows the same video to produce multiple overlapping clips as a type of data augmentation, increasing data variability, and improving alignment. TeachText \cite{ref_teachtext}, on the other hand, proposes a teacher-student distillation approach for Video-Text retrieval where, during training, the model uses knowledge of multiple text embeddings.

While most of the previously mentioned models are trained based on pre-trained text-image architectures like CLIP, our goal is to adopt models already pre-trained for video-text tasks, to further minimize data requirements. Specifically, we adopt X-CLIP in this paper. Our decision is based on the fact that X-CLIP is a large model available pre-trained under different configurations of number of frames, patch size, and fine-tune datasets. 

The X-CLIP architecture can be described in a simplified way as follows \cite{ref_clip}: as in most of the previously described models, it also contains Transformer-based text and vision encoders. However, it introduces a Multi-Frame Integration Transformer (MIT) and a Prompt Generator. A simplified sketch of the architecture with some modification is shown later in Fig.~\ref{fig:LoRA_fine_tune}. 

The vision Transformer is applied to each individual frame, extracting embeddings $\mathbf{f}$ from them: $\mathbf {F} = [\mathbf{f}_{1}, ..., \mathbf {f}_{j}, ..., \mathbf {f}_{T}]$ \cite{ref_xclip}, being $\mathbf {f}_{j}$ the vector representation of frame $j$ and $T$ the total number of frames. Then, the MIT transformer block receives $\mathbf {F}$ as input, performing self-attention between the frame vectors, allowing the model to understand the relationship between frames. Finally, the textual vectors produced by the text encoder and the resulting vectors from the MIT block serve as input to the prompt generator, which contains a cross-attention mechanism. The intuition behind the Prompt Generator proposed by the X-CLIP authors is that it acts like a text decoder that enhances the input text with information coming from the video.

\subsection{Fine-Tuning} 
\label{subsec:fine_tuning}

As previously mentioned, some of the vision-text models are trained by leveraging pre-trained Image-Text models. When explicitly speaking about fine-tuning vision models, two common parameter-efficient approaches usually appear in the literature: Prompt Tuning and Adapter Tuning. Prompt Tuning methods introduce learnable parameters at the input of  transformer layers (prompt or token level), while Adapter Tuning add parameters (or adapters) into the actual model layers. Multi-grained Prompt Tuning (MPT)~\cite{ref_mpt} is an example of Prompt Tuning method for video-text models, where the authors propose a video encoder composed of mainly three prompts: a spatial prompt, a temporal prompt, and a global prompt that aims to capture different characteristics of the video. RAP \cite{ref_rap}, on the other hand, is an Adapter Tuning example, where a video-text retrieval model is created by fine-tuning CLIP. 

An Adapter Tuning method that has been widely popularized in the context of Large Language Models but is little explored in the context of vision models is Low-Rank Adaptation (LoRA) \cite{ref_LoRA}. The authors in \cite{ref_LoRA_vision} compare LoRA with several other fine-tuning methods in the context of few-shot learning and demonstrate that LoRA outperforms competing approaches. Inspired by this work, and given that LoRA is easily accessible in libraries such as HuggingFace's PEFT \footnote{https://huggingface.co/docs/peft/}, we adopted LoRA as our fine-tuning method in this paper.

\subsection{Large Language Models and Agents} 
\label{subsec:llms_and_agents}
In previous sections, we detailed Vision-Language Models (VLMs) pre-trained with contrastive objectives, such as CLIP, and some of their adaptations to the video domain. Another key class of models that has recently gained attention is Large Language Models (LLMs). Like most VLMs, LLMs are also transformer-based architectures, but containing billions of parameters and trained on massive text datasets to perform text generation and understanding \cite{ref_llms_survey}.

The improvement of both VLMs and LLMs has led to the development of Multimodal Large Language Models (MLLMs) \cite{ref_llms_survey}, which combine vision and language capabilities, or even other modalities such as audio. As described in \cite{ref_llms_survey}, MLLMs are typically built integrating pre-trained modality encoders, such as CLIP, with LLMs, using modality connectors to align features across different domains. Several MLLMs are now available, including proprietary models such as the OpenAI's GPT, Google's Gemini, and Anthropic's Claude model families, as well as several open-source alternatives like LLaMA \cite{ref_LLaMA} and BLIP family \cite{ref_blip1}, \cite{ref_blip2}, \cite{ref_blip3}, to name a few. Due to their scale and to the large amount of training data, these models excel in zero-shot and few-shot learning tasks \cite{ref_llms_survey}, where they are able to perform well on unseen inputs with no (zero-shot) or minimal (few-shot) task-specific examples.

The ability of these models to generalize to unseen domains is, of course, limited by the number of parameters and by the quality and size of the training datasets. In order to further increase the capabilities of LLMs, several approaches have been proposed, such as Chain of Thoughts (CoT) prompting \cite{ref_llms_survey}. However, a particularly promising, recent and widely adopted method is the use of LLM-based agents \cite{ref_llm_agents}. LLM agents are autonomous and isolated systems that use LLMs to perform well-defined tasks. Usually, they are systems that use LLMs to plan and reason about the sequence of tasks that must be done to accomplish a particular and specific goal. To perform tasks, the LLMs are given access to tools, such as functions, Application Programming Interfaces (APIs), and databases, that extend their capabilities beyond the knowledge they have from the training data. Moreover, agents usually have access to memory, allowing the model to observe and remember the result of actions and plan its future steps \cite{ref_llm_agents}.

To the best of our knowledge, no existing work uses LLM agents to remove misaligned video-text pairs, as we do in this paper and detail in Section \ref{sec:data_treatment}. However, several studies explore video-text asymmetry and misalignment using pre-trained VLMs. For example, the authors in \cite{ref_info_asymmetry} use BLIP-2 and GPT-4 as image captioners to enrich and augment datasets at both the training and retrieval stages. In-Style \cite{ref_in_style} begins with unmatched texts and applies pseudo-matching using pre-trained image-text models like CLIP. A captioner model such as BLIP is then trained on the generated pairs to adapt query styles to videos, and these aligned pairs are used to fine-tune a video-text dual encoder. Other related works include "Text Is MASS" \cite{ref_text_is_mass} and \cite{ref_noisy_pair_calibration}.

\subsection{Animal Behavior Detection}
%In this section, we highlight other studies that, as ours, use Machine Learning with the objective of understanding animal behaviors. 
Understanding of animal behavior is explored in works like ChimpVLM \cite{ref_chimpvlm}, where a VLM is created to classify chimpanzee behaviors using the PanAf500 and PanAf20K datasets \cite{ref_panaf20k}. As in our case, the authors also do not rely solely on classification labels. Instead, they initialize query tokens using an ethogram of chimpanzee behaviors. DeepEthogram \cite{ref_deepethogram} employs pre-trained Convolutional Neural Networks (CNNs) on large open-source datasets to extract features from single video frames and classify them into user-defined behaviors. Finally, the study in \cite{ref_automated_audiovisual} develops an automated pipeline to distinguish two specific behaviors of chimpanzees from raw video data: buttress drumming and nut cracking. This pipeline integrates audio and video frame extraction, body tracking via CNNs, and behavior detection.

Additionally, Animal-Bench \cite{ref_animal_bench} and MammalNet \cite{ref_mammalnet} introduce benchmark datasets, while \cite{ref_zero_shot_animal} and \cite{ref_video_foundational_animal} evaluate and benchmark pre-trained Vision-Language foundational models for behavior analysis tasks such as classification.

Unlike the previously described works, in this paper we fine-tune a model using unlabeled raw video footage. Through our data processing pipeline, we extract informative text-video pairs to train the model. Moreover, rather than introducing a new architecture, we fine-tune a well-established pre-trained video-text model using Low-Rank Adaptation.

\section{Method}
\label{sec:method}

In this section, we describe our method in detail. Our goal is to be able to fine-tune a video-text model using our domain data and based solely on the raw videos we have available, \textit{i.e.}, without relying on manual labels, which are scarce and hard to obtain. As previously mentioned, what makes the problem especially challenging is the fact that the video quality is poor and, in particular, the fact that most of the audio descriptions are useless. Although the field collaborators sometimes describe the observed monkey behaviors, most of the audios are about unrelated subjects, past observations, or even about future intentions of the collaborators. Prediction errors in the adopted audio-text model can also produce degraded transcripts.

To handle such a complex problem, we intensively rely on the power of pre-trained models. Our approach is two folded. First, we propose a data treatment pipeline that uses Whisper \cite{ref_whisper}, BLIP-2 and LLaMA to filter video-text pairs into a subset of informative data. Implementation details are described in Section~\ref{sec:data_treatment}. In the second part, we fine-tune a pre-trained X-CLIP model using Low-Rank Adaptation (LoRA), as described in Section~\ref{sec:xclip_fine_tuning}.

\begin{figure*}[ht]
  \centering
  \begin{subfigure}[b]{0.48\textwidth}
    \caption{\textbf{Example of noisy clip-transcript pair. Transcript}: "Let's try to get closer to the female monkey to try to catch that interaction."}
    \includegraphics[width=\textwidth]{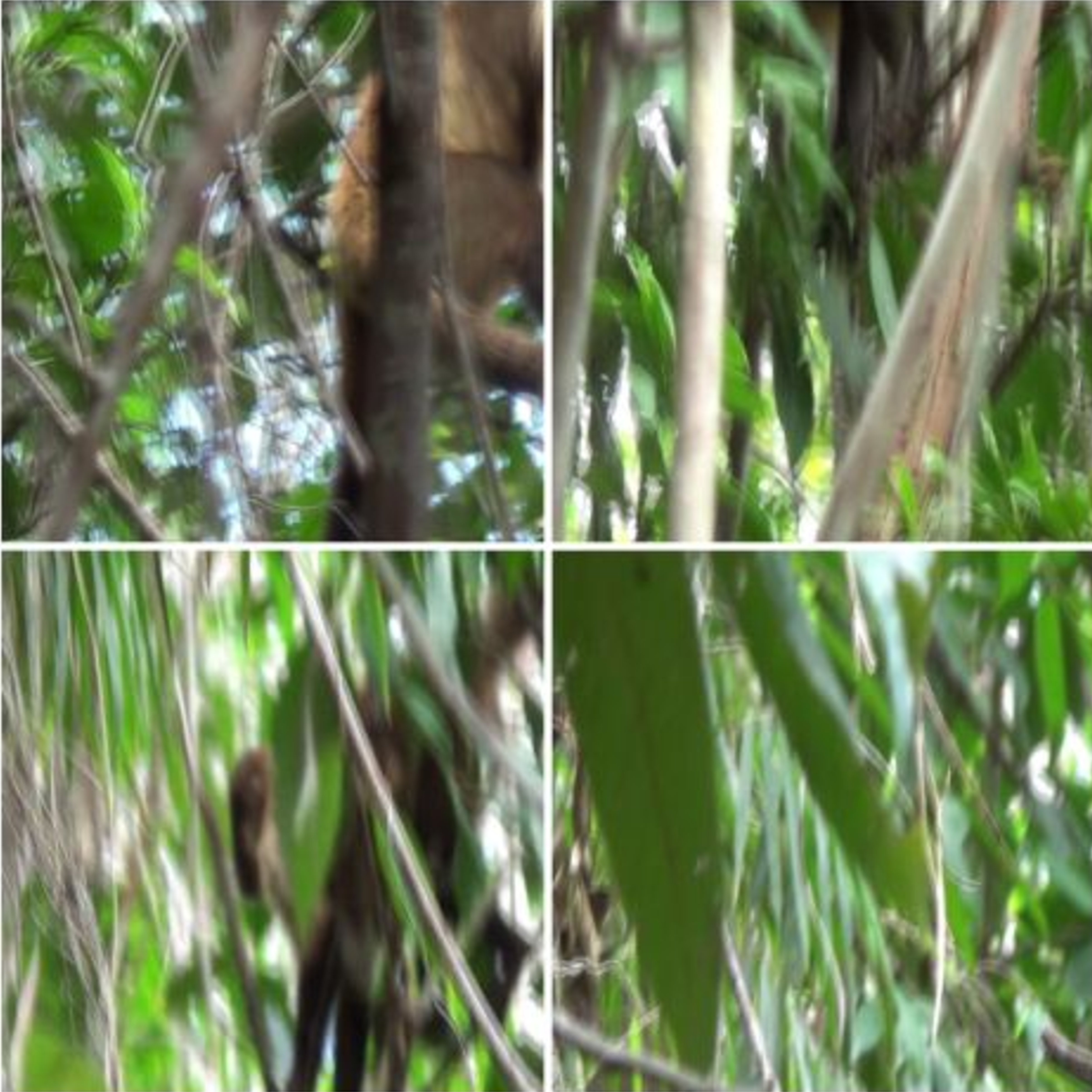}
  \end{subfigure}
  \hfill
  \begin{subfigure}[b]{0.48\textwidth}
    \caption{\textbf{Example of informative clip-transcript pair. Transcript}: "The hug is happening again between monkeys."}
    \includegraphics[width=\textwidth]{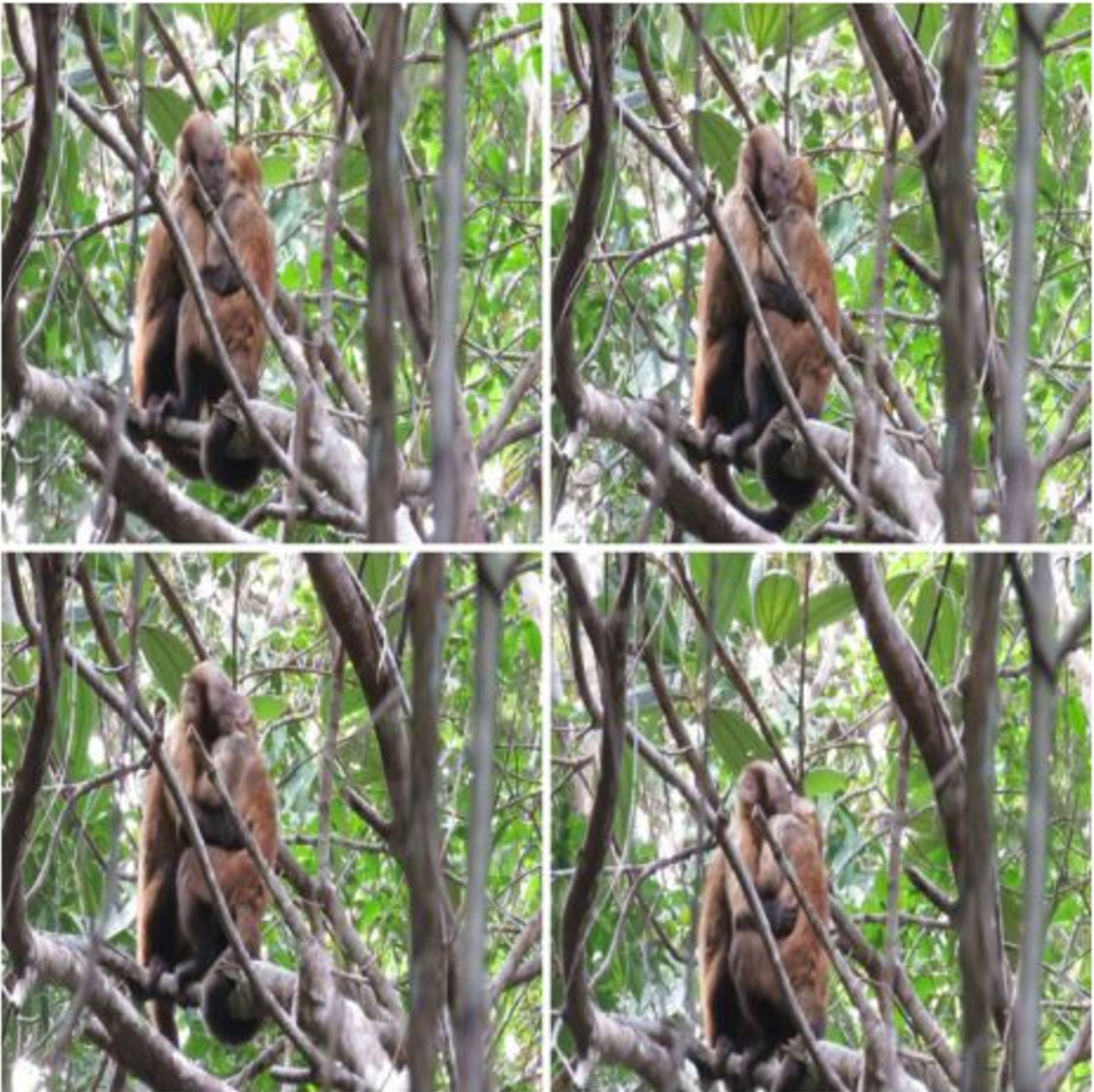}
  \end{subfigure}
  \caption{The figure presents two examples of video clips. In (a), we illustrate a commonly encountered video clip where the transcript poorly matches the content and the video itself is highly noisy. In (b), we display a video clip obtained after applying the proposed agentic data processing pipeline, resulting in significantly improved video quality and accurate alignment with its transcript. Copyright © LEDIS-USP archive.}
  \label{fig:good_vs_bad_images}
\end{figure*}

To illustrate the type of data we are working with, in the left (a) of Figure \ref{fig:good_vs_bad_images} we show a typical video-text pair found in the raw dataset: the camera may shake significantly, the monkeys are distant from the camera or between many vegetation, and the transcript is not informative. In the right (b) we see a video-text pair obtained after data processing, illustrating that useful information can be obtained from the raw dataset.

\subsection{Data Treatment Agent}
\label{sec:data_treatment}

\begin{figure*}[ht]
\centering
\includegraphics[width=\linewidth]{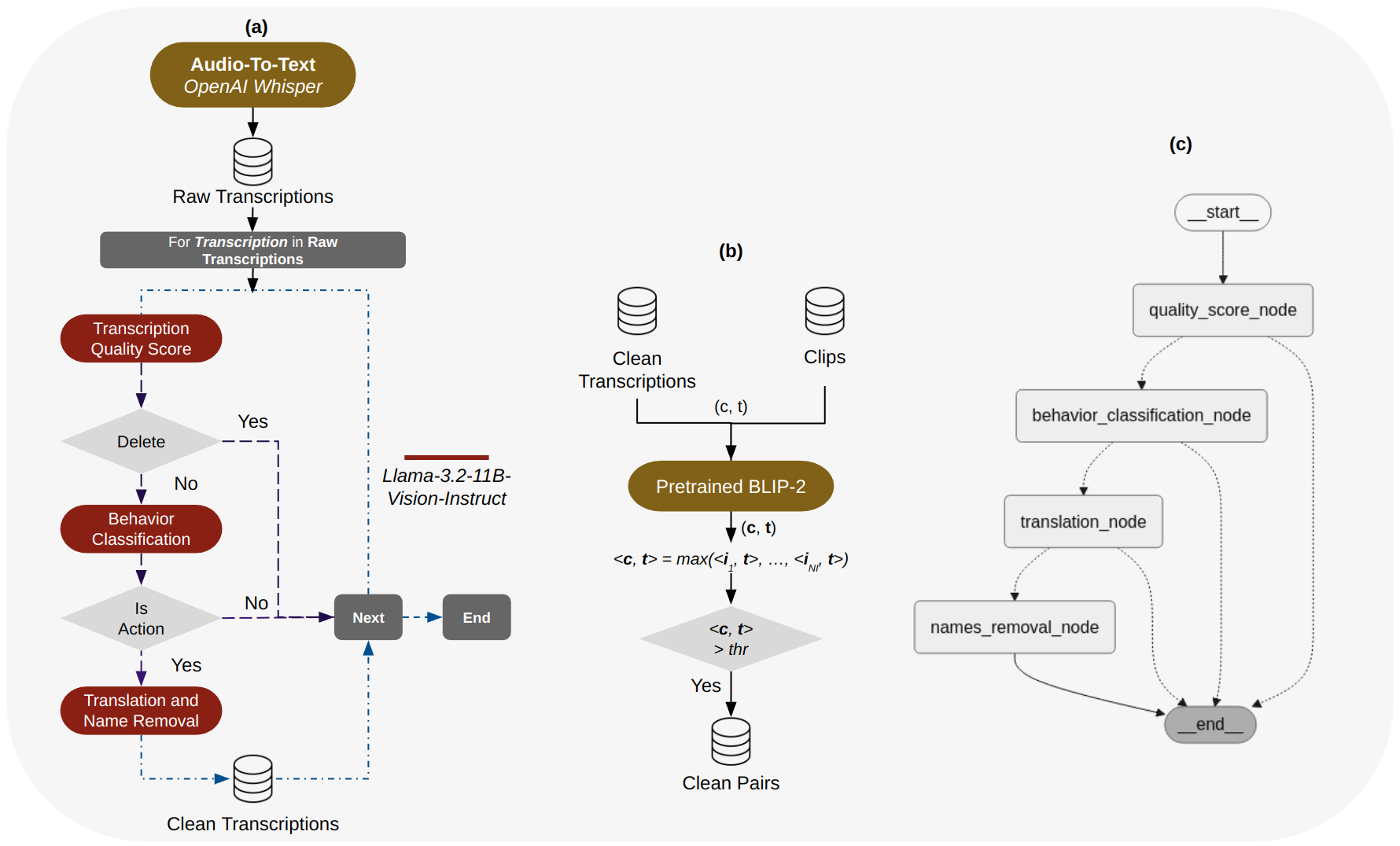}
\caption{\textbf{Data Generation Pipeline}. In (a) one can see that OpenAI's Whisper is used to extract raw transcripts, which are then treated by a data processing Agent. The clean (clip, transcript) pairs are then submitted into BLIP-2 model (b) and only the pairs with cosine similarity greater than a predefined threshold are maintained, reducing the amount of noisy pairs. The diagram in (c) shows the actual graph produced with LangGraph. The graph is applied to each raw transcript individually.}
\label{fig:data_clean_agent}
\end{figure*}

Inspired by the emergency of LLM-based agentic systems (Section~\ref{subsec:llms_and_agents}), we built a data processing pipeline to treat our dataset. Although our pipeline does not contain a reasoner, it can be understood as a simple agent that contains a chain of LLM-based processing steps that handles audio transcripts, as shown in Figure \ref{fig:data_clean_agent}.

More specifically, we adopt Meta's LLaMA 3.2 11B Vision model \footnote{meta-llama/Llama-3.2-11B-Vision-Instruct}, OpenAI's Whisper Large-V3 \footnote{https://github.com/openai/whisper} \cite{ref_whisper} and BLIP-2 \footnote{Salesforce/blip2-itm-vit-g-coco} contrastive module \cite{ref_blip2}. Our agent is built with LangGraph \footnote{https://github.com/langchain-ai/langgraph} and applies a sequence of the following tasks: convert the audio of the videos into text (\textbf{Transcription}); creates a \textbf{Quality Score} for the transcripts; classifies \textbf{Monkeys Behaviors}; \textbf{Translates} transcripts from Brazilian Portuguese to English and \textbf{Remove Names}. The clean transcripts obtained through this agent are then hard filtered using the BLIP-2 contrastive module to remove too noisy clip-transcript pairs. Notice that we used the pre-trained BLIP-2 version that is fine-tuned in Microsoft's COCO dataset, which have several samples containing animals \cite{ref_coco_dataset} and, therefore, is more suited to our domain. 

Each step of the pipeline is detailed in the following sections. After processing, only the clip-transcript pairs that meet all cleaning criteria are retained, ensuring higher quality and better alignment.

\subsubsection{Audio Transcription}
\label{subsubsec:audio_transcription}

%The video clip descriptions come from the audio transcriptions and we use OpenAI's Whisper Large-V3-Turbo \cite{ref_whisper} for that. 

To generate the transcription, we apply OpenAI's Whisper Large-V3-Turbo model~\cite{ref_whisper}. Whisper generates pairs $(t, \, ts)$ where $t$ is a textual transcript and $ts =(t_{init}, t_{end})$ is the timestamp of the audio segment corresponding to the transcript. Then, from each $(t, \, ts)$ pair, using the timestamp we identify the video segment and build the clip-transcript pair $(c,t)$.

Although Whisper model supports translation, we transcribe the audio in its original language, which is Brazilian Portuguese. Translation is done in a later step of the pipeline using LLaMA 3.2 as detailed later.

\subsubsection{Quality Score}
\label{subsubsec:quality_score}

We use LLaMA-3.2 to calculate a binary quality score for each transcript, determining whether it should be immediately discarded. Many transcripts are irrelevant to analyze capuchin monkeys behavior or do not align with the corresponding video. These include instances where the field collaborators discuss their plans, explain why a recording was unsuccessful, describe observations not captured on video, or provide vague mentions of individuals. Real examples of such transcripts are the following. 
\begin{itemize}
    \item "We shall try to observe something."
    \item "Let us see if she will interact with the infant."
    \item "We will go after the monkey."
    \item "We shall keep an eye out to see if we observe something."
\end{itemize}

To handle such cases, we provide LLaMA 3.2 with a prompt that instructs it to evaluate if the transcript captures relevant content related to capuchin monkeys behavior. In case a relevant transcript is detected, the model returns a quality score of 1 and the transcript goes into the following step of the pipeline, otherwise the $(c, t)$ pair is removed.

\subsubsection{Behavior Classification}
\label{subsubsec:behavior_classification}

We design a prompt such that LLaMA 3.2 can classify if there is at least one monkey behavior associated to each transcript, based solely on its text. The behaviors considered are those described in the Ethogram in \textbf{Table \ref{tab:monkeys_behavior}}. In the prompt we also ask the model to provide a binary classification score about whether one of the listed behaviors can be accurately detected through the transcript or not. In case a behavior cannot be detected for a given transcript, it is discarded.

\begin{table*}
\scriptsize
\caption{\textbf{Adopted Ethogram}. This is the ethogram adopted in this paper, which is adapted from \cite{ref_capuchin_monkeys_thesis}. We use the actions and descriptions here provided in the data treatment pipeline.}
\label{tab:monkeys_behavior}
\begin{tabularx}{\textwidth}{VZ}
\hline 
\textbf{Action} & \textbf{Description} \\ \hline \hline  Forage & Searches for food. \\
Predation & Attempts to capture prey.\\
Eat & Chews and swallows food. \\
Sample & Sniffs or bites food without eating. \\
Stand Still & Remains motionless while awake. \\
Rest/Sleep & Rests sitting or lying down. \\
Move, Walk or Run & Moves using all four limbs. \\
Bipedal Action & Moves or stands on two feet. \\
Locomotion While Foraging & Carries food while moving. \\
Grooming & Cleans another monkey’s fur. \\
Touch & Places hand on another monkey. \\
Nurse & Feeds from female’s breast. \\
Rest in Group & Rests in contact with others. \\
Play & Engages in non-aggressive play. \\
Lipsmack & Rapidly presses and opens lips. \\
Sexual & Mounting, body touching, genital contact, or copulation. \\
Scrounge & Collects and eats dropped food. \\
Beg Food & Requests food using gestures. \\
Alocarrying & Carries another monkey on its back. \\
Hug & Embraces another monkey. \\
Threatening & Displays aggressive facial expressions. \\
Double Threatening & Two monkeys threaten simultaneously. \\
Chase & Pursues another monkey. \\
Fight & Engages in violent conflict. \\
Vigilant & Scans surroundings with raised head. \\
Runaway & Moves away from a threat. \\
Sexual Self-Inspection & Manipulates own genitals. \\
Anointing & Rubs chewed substances on fur. \\
Urine Washing & Rubs urine on its own body. \\
Autoplay & Plays alone. \\
Auto-Grooming & Grooms itself. \\
Scratch & Rubs to relieve itching. \\
Yawn & Opens mouth wide and breathes deeply. \\
Nose Wipe & Touches own nose. \\
\hline
\end{tabularx}
\end{table*}

\subsubsection{Translation}
\label{subsubsec:translation}

We use LLaMA 3.2 to translate transcripts from Brazilian Portuguese to English. Translation is a necessary step because most pre-trained multimodal models are predominantly trained with English text. We prefer LLaMA 3.2 for this task instead of directly using Whisper because it allows us to customize the prompt with a glossary for specific terms that might confuse the translator. For example, the Portuguese phrase "está rolando uma interação entre os indivíduos" is an informal way of saying "the individuals are interacting." However, the word "rolando" also means "rolling," which could mistakenly suggest a monkey behavior instead of an interaction. After translation, we also ask the model to replace any monkey name by the words "monkey" or "capuchin monkey".

\subsubsection{Noise Filtering}
\label{subsubsec:noise_filtering}

Finally, in the last filtering step, we use pre-trained Salesforce's BLIP-2 \cite{ref_blip2} fine-tuned on Microsoft's COCO dataset to compute the cosine similarity between the embeddings of the clip and the transcript for every $(c, t)$ pair and remove the pairs that have low similarity. We decided to use an image-text model instead of X-CLIP itself to not bias the fine-tuning process. 

The way we use an image-text model to filter clip-transcript pairs is the following: since clip $c$ can also be understood as a sequence of images $c = (i_{i}, i_{2}, ..., i_{N_{I}})$,
for a given clip-transcript pair $(c,t)$
we consider instead the pair $((i_{i}, i_{2}, ..., i_{N_{I}}), t)$. To compute the similarity between the sequence of images and the transcript, we first use BLIP-2 to obtain the images and transcript embeddings $\mathbf {i} \in \mathbb{R}^{p}$ and $\mathbf {t} \in \mathbb{R}^{p}$. Then we compute the maximum among the cosine similarity of each image-transcript pair: $\langle\mathbf{c}, \mathbf{t}\rangle = max(\langle\mathbf{i}_{1}, \mathbf{t}\rangle, \langle\mathbf{i}_{2}, \mathbf{t}\rangle, ..., \langle\mathbf{i}_{N_{I}}, \mathbf{t}\rangle)$, as shown in Figure~\ref{fig:data_clean_agent}(b). Finally, we remove the pairs with $\langle\mathbf{c}, \mathbf{t}\rangle$ below $0.32$. This threshold was obtained through visual inspection: we tried several threshold values and checked the retrieval results. We then selected a value below which the images were clearly absolute noisy and uncorrelated with the transcripts.

\subsection{X-Clip Fine-Tuning}
\label{sec:xclip_fine_tuning}

\begin{figure}[ht]
\centering
\includegraphics[scale=0.08]{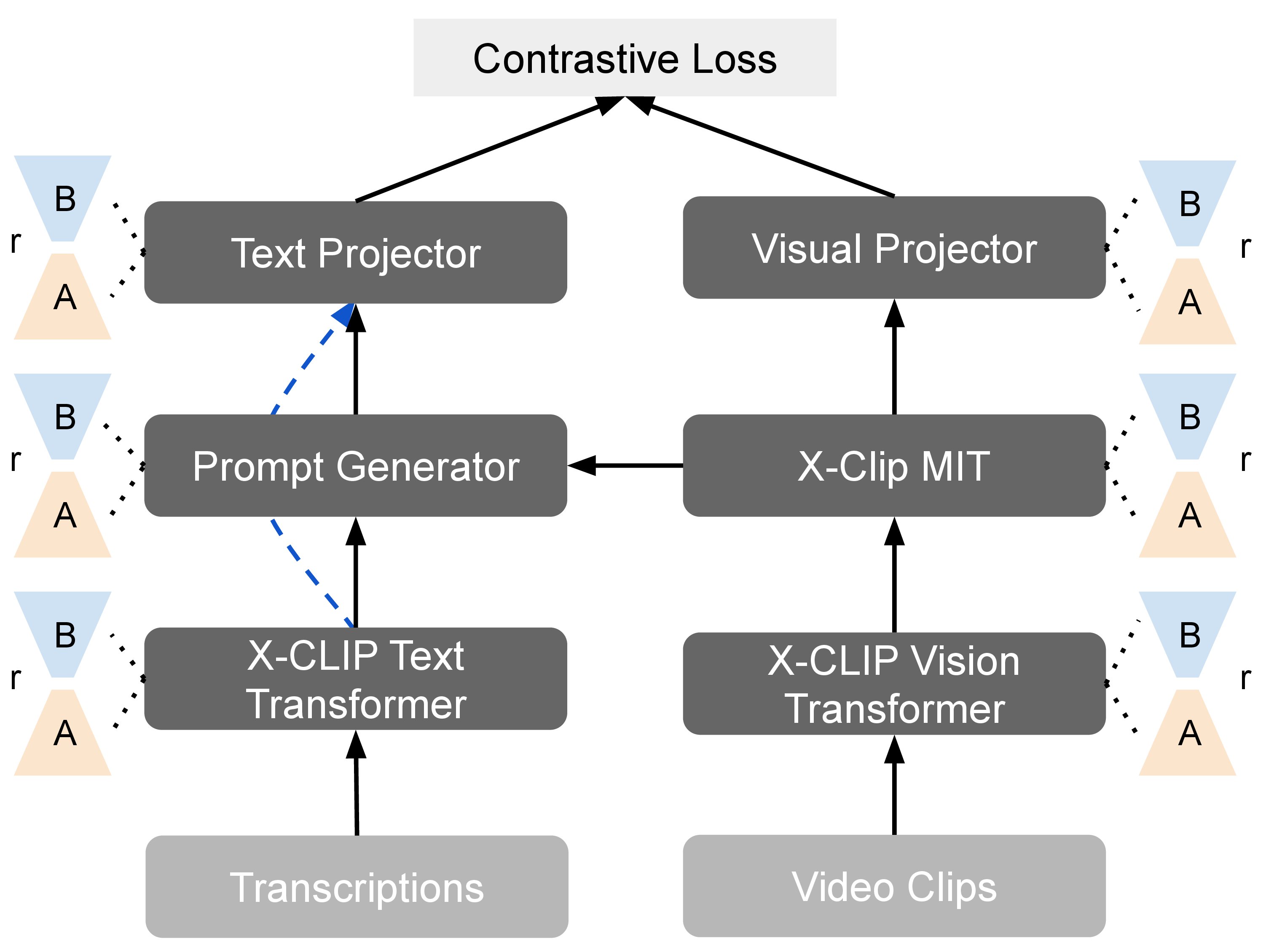}
\caption{\textbf{X-CLIP Fine-Tuning Architecture.} The X-CLIP architecture is the one proposed in \cite{ref_xclip} and here simplified. It consists of a text and a vision transformer, a Multi-Frame Integration Transformer (MIT), a Prompt Generator and projections to map both modalities into the same vector space. We include LoRA layers in all of the mentioned blocks. The blue dashed arrow represents the additional textual embedding that is incorporated into the loss function, generated bypassing the Prompt Generator and feeding the output of the Text Transformer directly into the Text Projector. The horizontal arrow from the MIT block to the Prompt Generator highlights that the Prompt Generator receives both text and vision inputs.}
\label{fig:LoRA_fine_tune}
\end{figure}

The second step of our method is to fine-tune X-CLIP model with the data obtained in Step 1 (Section \ref{sec:data_treatment}). To adapt X-CLIP to our domain, we include LoRA layers in all the transformer blocks of the model, as well as in the final projection layers, as shown in Figure \ref{fig:LoRA_fine_tune}. LoRA \cite{ref_LoRA} is a fine-tuning method that was popularized in the context of Large-Language Models and few works have addressed its use in the context of vision models. In-depth analyses on the impact of LoRA on vision models are given in \cite{ref_LoRA_vision}, where the authors show that LoRA outperforms most of the fine-tuning techniques they analyzed. Given a pre-trained layer $\mathbf {W} \in \mathbb{R}^{p \times q}$ \cite{ref_LoRA}, LoRA constrains the update of $\mathbf {W}$ through low-rank decomposition: $\mathbf {W} + \mathbf {B}\mathbf {A}$, where  $\mathbf {B} \in \mathbb{R}^{p \times r}$ and $\mathbf {A} \in \mathbb{R}^{r \times q}$ are learnable linear matrices \cite{ref_LoRA}.

Notice in Figure \ref{fig:LoRA_fine_tune} that there is a Prompt Generator layer in the path that generates the Text Embedding. As explained in Section~\ref{subsubsec:existing_models}, the Prompt Generator is a decoder that enriches the text embedding with video information through a cross-attention mechanism. This is particularly useful for Zero-shot classification, where the input texts are usually solely the class labels \cite{ref_xclip}.

However, our primary focus is video-text retrieval, not zero-shot classification, because our objective is to search for arbitrary monkey behaviors in videos. Therefore, the cross-attention mechanism introduces a challenge: it requires simultaneous access to both video and text inputs, preventing independent computation of their embeddings. This is problematic for retrieval, where we typically compute and store vision embeddings in a vector database for later querying.

To enable retrieval, we could, instead, use the independently computable embeddings from the Text and Vision Transformers (first module of each path) in the X-CLIP schema shown in Figure \ref{fig:LoRA_fine_tune}. However, these are not the embeddings directly optimized by the contrastive loss. To address this issue, we include one more component in the loss function, as given in Equation~(\ref{eq:proposed_loss}). The second term, $CL(c, t)$, directly uses the output of the projector. The first term, $\overline{CL}(c, t)$, on the other hand, adopts the text embedding produced by skipping the prompt generator, represented by the blue dashed line in Figure~\ref{fig:LoRA_fine_tune}. This embedding can be computed independently from the vision embedding, which is useful for retrieval. 

\begin{equation}
    \label{eq:proposed_loss}
    L = \frac{\overline{CL}(c, t) + CL(c, t)}{2}
\end{equation}

With this modification, we can then use the model in the following way:

\begin{itemize}
    \item For Zero-shot classification tasks, we just use both projector outputs;
    \item For retrieval tasks, we use the independently computable embeddings mentioned above.
\end{itemize}

\section{Results}
\label{sec:results}

In this section, we first introduce the dataset and the result of the data treatment pipeline described in Section \ref{sec:data_treatment}. Next we describe the adopted fine-tuning experimental setup and fine-tuning results for ranking (retrieval) and zero-shot classification tasks. Finally, for each task, we compare the selected best fine-tuned models with several raw pre-trained X-CLIP models, together with qualitative analyses. This last part aims to showcase how effective is the small yet carefully crafted fine-tuning dataset.

%The results show that the data processing followed by fine-tuning significantly improves performance on a manually labeled high-quality out-of-sample test dataset, despite training exclusively on highly noisy videos. We also display qualitative analyses that support the numerical results and analyze the impact of the LoRA rank and the selected fine-tuned model layers on the final performance metrics. \nina{será que isso não seria melhor na discussão ou na conclusão?}

\subsection{Data Treatment Pipeline Results}
\label{sec:dataset}
The dataset adopted in this paper is a sample of videos from the Una Biological Reserve (ReBio), located at state of Bahia, Brazil ($15^{\circ}6'-12'\,\text{S}$ and $39^{\circ}02'-12'\,\text{W}$) \cite{ref_capuchin_monkeys_thesis}. The data we are using is composed of six individuals (capuchin monkeys), with video recordings spanning from birth to 36 months of age. Moreover, the dataset is around 1.5 TB in size, with a total of 13,060 videos and 284 hours. After applying the transcription step described in Section \ref{subsubsec:audio_transcription}, the number of raw clip-transcript pairs obtained is 123,871. This number is reduced to only 7,862 clean pairs after applying the cleaning agent described in Section \ref{sec:data_treatment}, which belong to 4,764 distinct videos (161 hours).

\begin{figure}[ht]
\centering
\includegraphics[width=0.8\linewidth]{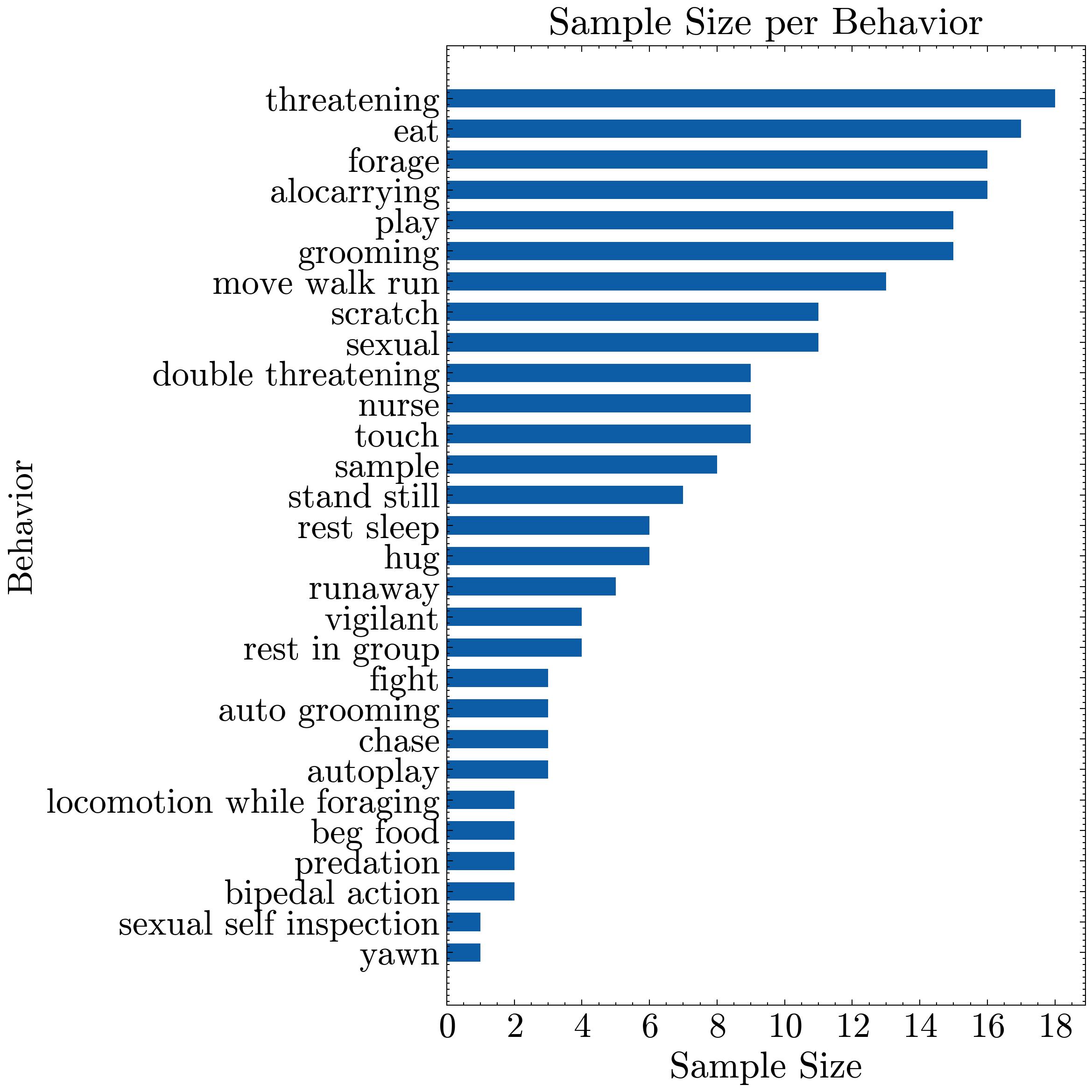}
\caption{Test dataset label distribution: number of instances per behavior.}
\label{fig:test_data_label_distribution}
\end{figure}

We then created a small test dataset with 177 clip-transcript pairs to be used as the out-of-sample test set. Each of these pairs is composed of a video clip and a corresponding manually annotated description of the clip, plus a list of the behavior types according to the ethogram in Table \ref{tab:monkeys_behavior}. We relied on the labels created by LLaMA 3.2 in the agentic pipeline to select an even number of instances per behavior type to be included in the test dataset, since some behaviors are really hard to find. Figure \ref{fig:test_data_label_distribution} shows the distribution of instances per behavior in the test set. The 177 clip-transcript pairs in the golden dataset are drawn from 166 unique videos, all of which are excluded from the training data.

Finally, it should be mentioned that the video clips generated from the raw footage were obtained through sparse sampling, which means that frames were sampled at regular intervals to efficiently capture relevant information while minimizing redundancy. We produced video clips with 8 and 16 frames, which are the amount of frames supported by X-CLIP pre-trained models.

\subsection{Fine-Tuning}
\label{sec:experimental_setup}

With regard to LoRA based fine-tuning of Vision-text models, the authors in \cite{ref_LoRA_vision} show that adapting $\mathbf{W_{v}}$ (value) and $\mathbf{W_{o}}$ (output) attention matrices produced better fine-tuning results on average. The authors also showed the impact of the location of the LoRA modules. 
Drawing inspiration from their work, our experiments are conducted as follows. 

We include LoRA layers in both projectors (which are linear layers), in both $\mathbf{W_{v}}$ and $\mathbf{W_{o}}$ matrices, and in the corresponding feed-forward layers that proceed $\mathbf{W_{o}}$. Both matrices exist in every multi-head self-attention module, but we don't fine-tune all of them. Our choices are the ones below:

\begin{itemize}
    \item The MIT contains a single encoder, while Prompt Generator contains two decoders. We only fine-tune the upper (second) decoder of Prompt Generator, because fine tuning bottom layer increases memory requirements;

    \item X-CLIP vision and text Transformers contain 12 encoders. Therefore, we test including LoRA in three different ways:
    
    \textbf{Upper Layers}: we add LoRA layers only in the 11th and 12th encoders;

    \textbf{Bottom Layers}: we add LoRA layers only to the 6-th encoder;

    \textbf{Vertical Layers}: we add LoRA layers to the 6th, 9th and 12th encoders. 
\end{itemize}

We also experiment with the LoRA ranks, trying values of 1, 2, 4 and 8. The fine-tuning is applied to two X-CLIP base models: base-patch16-kinetics-600-16-frames \footnote{microsoft/xclip-base-patch16-kinetics-600-16-frames} for the 16 frames model, and base-patch16-kinetics-600 \footnote{microsoft/xclip-base-patch16-kinetics-600} for the 8 frames model. We choose these as the base models for fine-tuning because they are pre-trained on the Kinetics-600 dataset, a large-scale action recognition dataset of videos from 600 action categories.

\paragraph{\textbf{Training Parameters}} Fine-tuning is conducted using the AdamW optimizer with a weight decay of 0.8 and a dropout rate of 0.5 in the LoRA weights. A batch size of 8 is used with gradient accumulation over 10 steps, effectively simulating a batch size of 80. Gradient clipping is applied with a max norm of 1.0. The learning rate is scheduled via Cosine Annealing, with a linear warm-up over one epoch, reaching a peak of  $1 \times 10^{-3}$.

\paragraph{\textbf{Computational Resources}} Experiments were conducted on a local machine equipped with an NVIDIA RTX-4090 GPU (24 GB VRAM), 64 GB RAM, and an AMD Ryzen 7 5800X CPU (8 cores, 16 threads, up to 4.7 GHz).

\paragraph{\textbf{Fine-tuning Models}} Our primary interest is to evaluate the fine-tuned models on the retrieval task. In fact, the contrastive loss is designed aiming optimization for retrieval tasks. Nevertheless, we also evaluate the models on the zero-shot classification task, with respect to the class label (manually assigned behavior type listed in Figure \ref{fig:test_data_label_distribution}).

For the {\bf retrieval task}, for each transcript in the test set, we rank the clip embeddings according to their cosine similarity to the transcript embedding. If the clip corresponding to the transcript is among the top-K ranked ones, then it is a Top-K hit. For the {\bf zero-shot classification task}, instead of computing embeddings for the transcripts, the embeddings are computed for each of the class label texts. Then, for each clip we rank the class label embeddings according to the cosine similarity to the clip embedding, and verify whether the clip label is among the top-K ranked class labels.

Because we are testing different LoRA parameters and fine-tuning configurations, it is expected that the best model for retrieval differs from the best model for zero-shot classification, as these are fundamentally distinct tasks. Similarly, it is natural that the optimal model varies by Top-K: a model that excels at Top-1 accuracy may differ from one that excels at Top-5. This divergence arises because Top-1 focuses solely on identifying the single best guess, while Top-5 measures whether the correct answer appears among the top five, capturing a different aspect of performance. The key question, then, is which model is "better", something that depends on the specific goals of the application. In this paper, we adopt the following selection criterion: for both retrieval and zero-shot classification, we define the best model as the one achieving the highest average across Top-1, Top-2, and Top-3 accuracies, i.e., $\text{avg}(\text{Top-1}, \text{Top-2}, \text{Top-3})$. The best performing models are shown in Table~\ref{tab:best_models}. Notice that we evaluate both the 8 and 16 frames models.

\begin{table*}[ht]
\centering
\scriptsize
\caption{\textbf{Selected Models}. This table displays the best models selected for each individual task and by number of frames.}
\begin{tabularx}{\textwidth}{ZZZ}
\hline
\textbf{Best Model Name} & \textbf{LoRA Rank} & \textbf{Fine-Tuning Layers}
\\ \hline \hline  
LoRA-16-frames-retrieval & 4                  & Vertical                    \\
LoRA-16-frames-zero-shot & 8                  & Bottom                      \\
LoRA-8-frames-retrieval  & 8                  & Vertical                    \\
LoRA-8-frames-zero-shot  & 2                  & Bottom \\ \hline           
\end{tabularx}
\label{tab:best_models}
\end{table*}

Details on the impact of the LoRA parameters and fine-tuned layers in retrieval and classification metrics can be seen in \ref{appendix:LoRA_params}.

\subsection{Comparative and qualitative analysis}
\label{subsec:metrics}

%In this paper, we are more interested in retrieval capabilities, which is related to the model ability to rank items and to retrieve correct items (video clips) to the top-k results. Moreover, the contrastive loss is optimizing the retrieval task. However, we also decided to evaluate the zero-shot classification capability of the model, even fine-tuning it for retrieval. We evaluate zero-shot classification based on the manual labels produced for the out-of-sample dataset and described in Figure \ref{fig:test_data_label_distribution}.

In this section, we compare the selected models  with other pre-trained X-CLIP models on retrieval and zero-shot classification tasks. For the retrieval task, we show examples of clips retrieved based on textual descriptions. Evaluation metrics are all computed on the test set.

\subsubsection{Retrieval}
\label{subsubsec:retrieval_metrics}

%Retrieval metrics are the most relevant ones to our domain, because our goal is to facilitate the discovery of general monkey behaviors through text, making future manual analysis more efficient. Therefore, 
We first evaluate the model using Hits@K. As shown in Table~\ref{tab:hits_at_k}, results from the pre-trained models highlight the difficulty of the task. None of them achieves 10\% on Hits@5 or 20\% on Hits@10. On the other hand, our fine-tuned model presents significantly improved hits. Compared to the best pre-trained raw model, the 16-frame model shows a $167\%$ improvement on Hits@5 and $145\%$ on Hits@10. For the 8-frame model, the gains are $114\%$ and $85\%$, respectively.

%Despite selecting high-quality videos for evaluation, challenges persist due to factors such as monkeys being distant from the camera, obscured by dense vegetation, or recorded under varying lighting conditions. 
%However, fine-tuning significantly improves the results: compared to the best pre-trained raw model, the 16-frame model shows a $167\%$ improvement in Hits@5 and $145\%$ in Hits@10. For the 8-frame model, the gains are $114\%$ and $85\%$, respectively.

\begin{table*}[ht]
\scriptsize
\centering
\caption{Hits@K of different pre-trained X-Clip and our pre-trained models}
\begin{tabularx}{\textwidth}{ZYY}
\hline
\textbf{Model}                            & \textbf{Hits@5}   & \textbf{Hits@10}  \\ \hline \hline
\multicolumn{3}{c}{\textbf{16 Frames}}                                            \\ \hline
base-patch16-16-frames              & 0.04          & 0.10          \\
large-patch14-16-frames             & 0.05          & 0.11          \\
base-patch32-16-frames              & \underline{0.06}          & \underline{0.11}          \\
base-patch16-kinetics-600-16-frames & 0.06          & 0.08          \\
\textbf{LoRA-16-frames-retrieval}             & \textbf{0.16} & \textbf{0.27} \\ \hline
\multicolumn{3}{c}{\textbf{8 Frames}}                                             \\ \hline
base-patch16-kinetics-600           & 0.03          & \underline{0.13}          \\
large-patch14-kinetics-600          & \underline{0.07}          & 0.11          \\
large-patch14                       & 0.06          & 0.11          \\
base-patch32                        & 0.02          & 0.09          \\
base-patch16                        & 0.07          & 0.10          \\
\textbf{LoRA-8-frames-retrieval}              & \textbf{0.15} & \textbf{0.24} \\ \hline
\textbf{Random Chance}                     & 0.03          & 0.06 \\ \hline    
\end{tabularx}
\label{tab:hits_at_k}
\end{table*}

We also evaluate the model ranking ability by using $NDCG@K$. This metric focus on class labels; for each class, it captures how many of the Top-K retrieved clips are true positives and if the positive ones are ranked above the negative ones. Table \ref{tab:ndcg_at_k} shows the $NDCG@5$ computed per class, considering only those with more than 10 samples, to ensure statistical reliability and to reduce noise from underrepresented classes. 

\begin{table*}[ht]
\scriptsize
\caption{NDCG@5 with respect to different behaviors: Threatening (\textbf{TH}), Sexual (\textbf{SX}), Forage (\textbf{FR}), Grooming (\textbf{GR}), Scratch (\textbf{SC}), Play (\textbf{PL}), Eat (\textbf{ET}), Move, Walk or Run (\textbf{MV}) and Alocarrying (\textbf{AC}).}
\begin{tabularx}{\textwidth}{ZWWWWWWWWW}
\hline
\multirow{2}{*}{\textbf{Model}}     & \multicolumn{9}{c}{\textbf{Behavior}}                                                                                                         \\
                                    & \textbf{TH}   & \textbf{SX}   & \textbf{FR}   & \textbf{GR}   & \textbf{SC}   & \textbf{PL}   & \textbf{ET}   & \textbf{MV}   & \textbf{AC}   \\ \hline \hline
\multicolumn{10}{c}{\textbf{16 Frames}}                                                                                                                                             \\ \hline
base-patch16-16-frames              & 0.00          & 0.00          & 0.06          & 0.00          & 0.00          & {\ul 0.07}    & {\ul 0.06}    & 0.08          & 0.06          \\
large-patch14-16-frames             & 0.00          & 0.00          & 0.06          & 0.00          & 0.00          & 0.00          & 0.00          & 0.08          & {\ul 0.13}    \\
base-patch32-16-frames              & 0.00          & 0.00          & {\ul 0.31}    & {\ul 0.07}    & 0.00          & 0.00          & 0.00          & 0.00          & {\ul 0.13}    \\
base-patch16-kinetics-600-16-frames & 0.00          & \textbf{0.18} & 0.13          & 0.00          & 0.00          & 0.00          & 0.00          & {\ul 0.15}    & 0.06          \\
\textbf{LoRA-16-frames-retrieval}             & \textbf{0.63} & 0.00          & \textbf{0.43} & \textbf{0.50} & 0.00          & \textbf{0.39} & \textbf{0.71} & \textbf{0.63} & \textbf{0.70} \\ \hline
\multicolumn{10}{c}{\textbf{8 Frames}}                                                                                                                                              \\ \hline
base-patch16-kinetics-600           & {\ul 0.11}    & 0.00          & 0.06          & 0.00          & 0.00          & 0.00          & 0.06          & {\ul 0.15}    & 0.06          \\
large-patch14-kinetics-600          & {\ul 0.11}    & 0.00          & {\ul 0.19}    & 0.00          & \textbf{0.18} & 0.00          & 0.12          & 0.00          & 0.06          \\
large-patch14                       & {\ul 0.11}    & 0.00          & 0.13          & 0.00          & 0.00          & 0.00          & 0.06          & 0.00          & {\ul 0.13}    \\
base-patch32                        & 0.00          & 0.00          & 0.06          & 0.00          & 0.00          & \textbf{0.07} & 0.00          & {\ul 0.15}    & {\ul 0.13}    \\
base-patch16                        & {\ul 0.11}    & 0.00          & 0.06          & 0.00          & 0.00          & 0.00          & {\ul 0.18}    & 0.08          & 0.06          \\
\textbf{LoRA-8-frames-retrieval}              & 0.00          & 0.00          & \textbf{0.85} & \textbf{0.63}          & 0.00          & 0.00          & \textbf{0.54} & \textbf{0.43} & \textbf{0.69} \\ \hline
\end{tabularx}
\label{tab:ndcg_at_k}
\end{table*}

As can be observed in Table \ref{tab:ndcg_at_k}, the raw models are in general completely unable to properly rank most of the behaviors considered. On the other hand, the fine-tuned versions of the model, especially the 16-frame version, present significantly better performance. Scratch and sexual behaviors appear as exceptions, which is comprehensible given that they are rare behaviors and are usually not directly described in detail by the field collaborators. 

\begin{figure*}[ht]
    \centering
    \begin{subfigure}{0.49\textwidth}
        \centering
        \includegraphics[width=\linewidth]{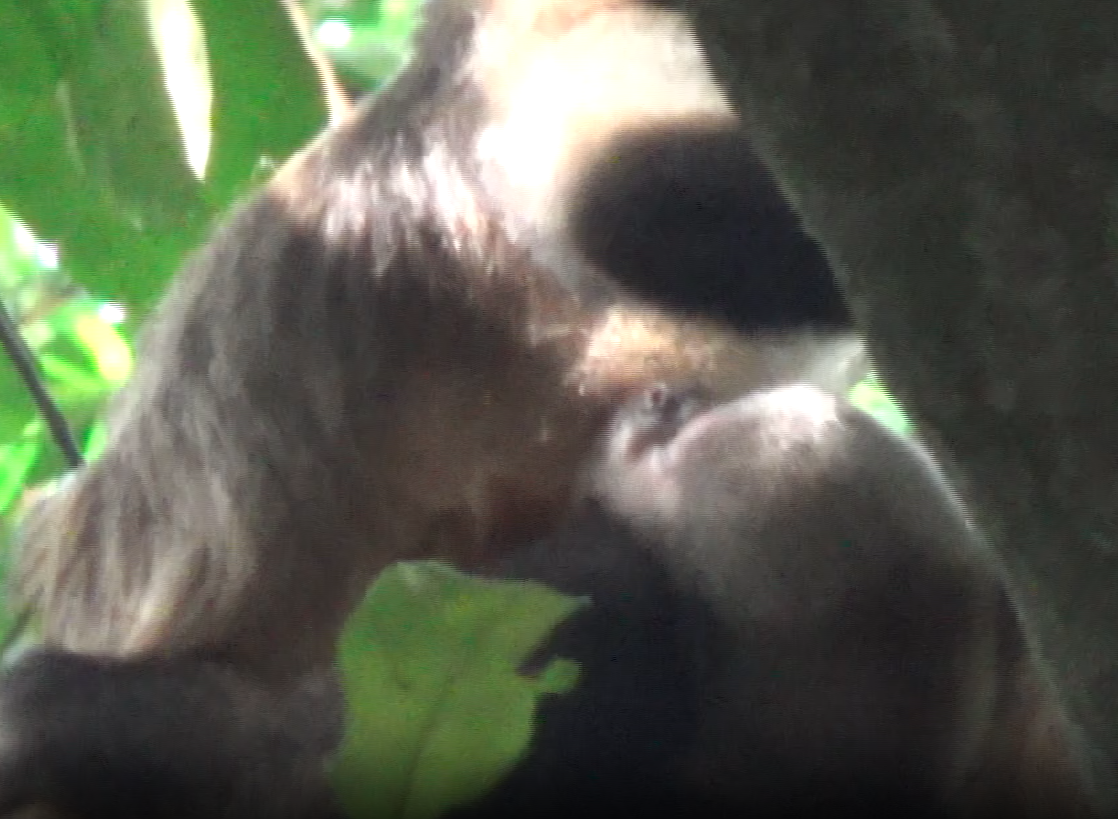}
        \caption{\textbf{Prompt:} "A young monkey nursing"}
        \label{fig:nursing}
    \end{subfigure}
    %\hspace{0.02\textwidth}
    \begin{subfigure}{0.49\textwidth}
        \centering
        \includegraphics[width=\linewidth]{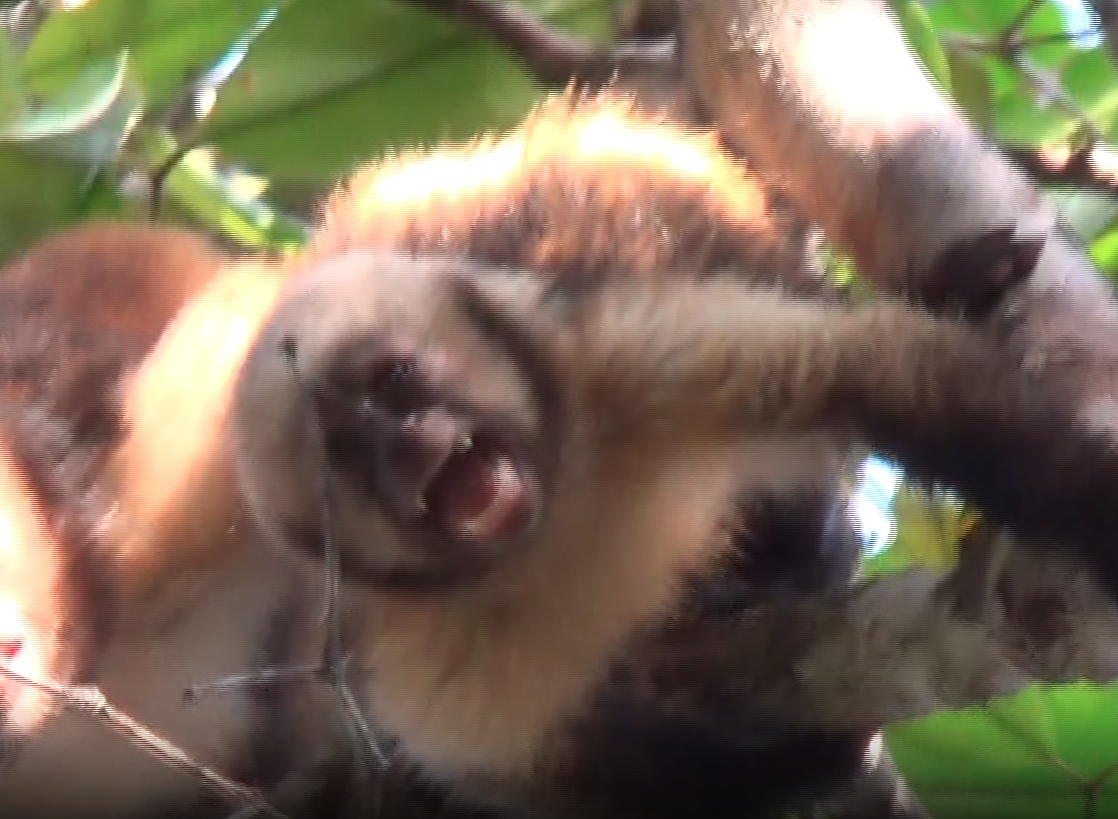}
        \caption{\textbf{Prompt:} "A monkey threatening something"}
        \label{fig:threatening}
    \end{subfigure}
    
    \vspace{0.1cm}
    
    \begin{subfigure}{0.49\textwidth}
        \centering
        \includegraphics[width=\linewidth]{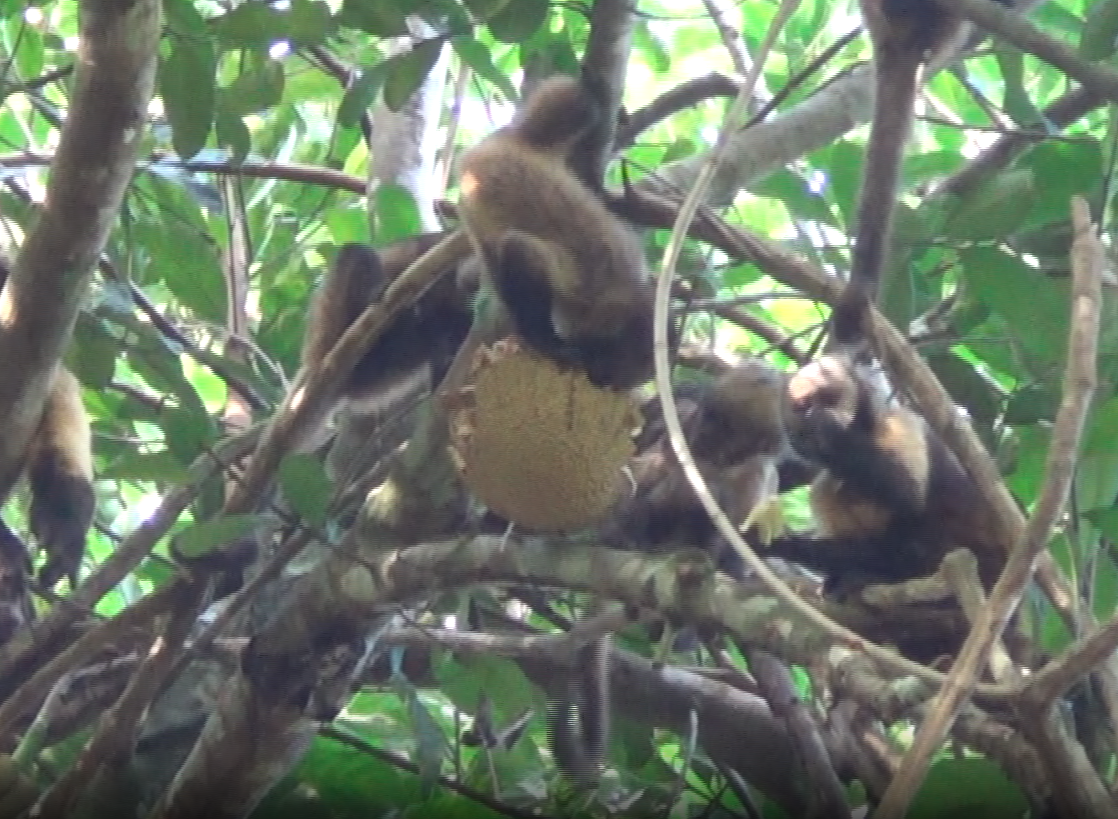}
        \caption{\textbf{Prompt:} "Monkeys eating a jackfruit"}
        \label{fig:jackfruit}
    \end{subfigure}
    %\hspace{0.02\textwidth}
    \begin{subfigure}{0.49\textwidth}
        \centering
        \includegraphics[width=\linewidth]{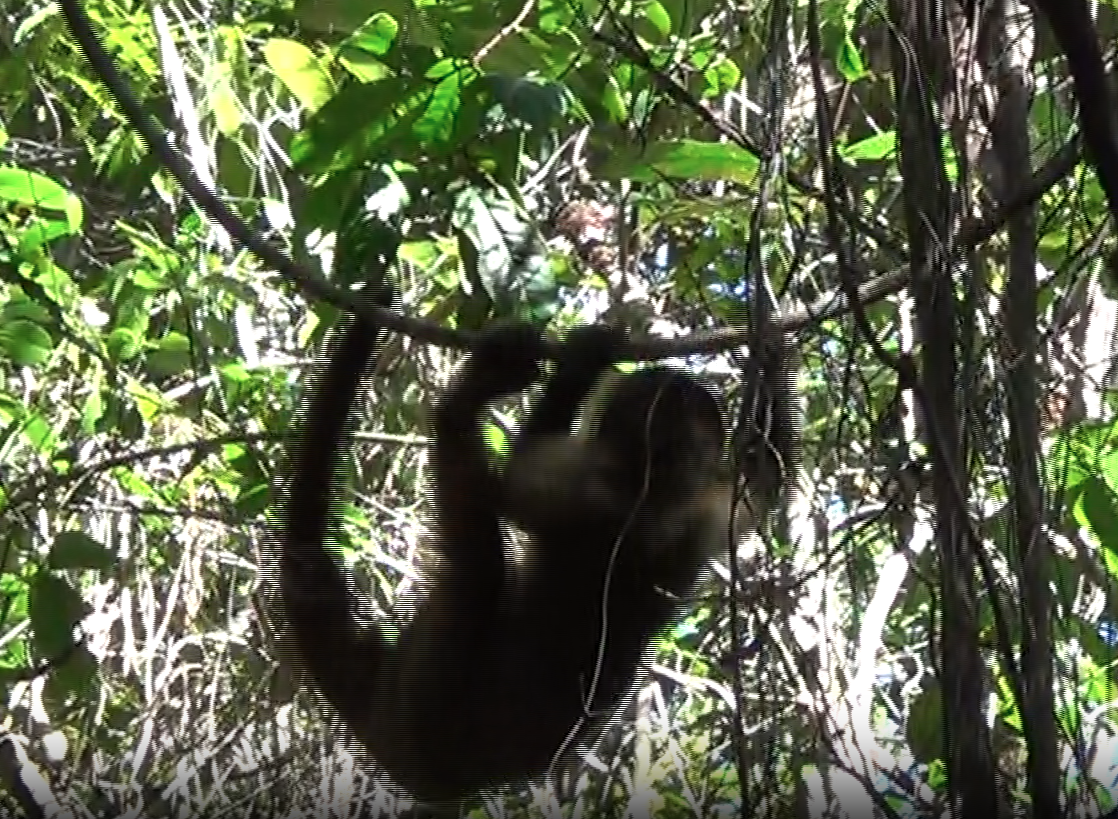}
        \caption{\textbf{Prompt:} "A monkey swinging on a vine"}
        \label{fig:swinging_on_vine}
    \end{subfigure}
    
    \caption{Resulting videos of Sapajus xanthosternos in Una Biological reserve for different input prompts. Copyright \textcopyright\ LEDIS-USP archive.}
    \label{fig:retrieval_clips}
\end{figure*}

To illustrate that the fine-tuned model is indeed able to generalize to our domain, we also present some qualitative examples. In particular, we input the model with the following prompts: "A young monkey nursing", "A monkey threatening something", "Monkeys eating a jackfruit" and "A monkey swinging on a vine". With those examples, we observe the model ability to capture specific behaviors such as nursing and threatening, present in the ethogram, as well as more generic behaviors such as "swinging on a vine". In Figure \ref{fig:retrieval_clips} we display the resulting video clips extracted from the test set for each of the four input prompts.

\subsubsection{Zero-Shot Classification}
\label{subsubsec:zero_shot_metrics}

%In Zero-shot classification, we provide the model with a video clip and the list of possible textual labels and we evaluate if the correct label appears in the Top-k predictions of highest probability. The labels we use are the ones in Figure \ref{fig:test_data_label_distribution}, which correspond to a total of 29 labels. Results can be seen in Table \ref{tab:zero_shot_metrics}. 
Table~\ref{tab:zero_shot_metrics} shows the zero-shot classification results. As one can observe, in this task the relative improvements are lower compared to the retrieval metrics. This may be justified based on the fact that the models were fine-tuned for the retrieval task. Nevertheless, compared to the best raw pre-trained models, for the 16 frames models we observe an improvement of $50\%$, $55\%$ and $13\%$ for Top-1, Top-5 and Top-10 accuracy, respectively; and, for the 8 frames model, an improvement of $100\%$, $29\%$ and $5\%$, respectively.

\begin{table*}[ht]
\centering
\scriptsize
\caption{\textbf{Zero-shot Accuracy}. Average accuracy for Top-1, Top-5 and Top-10 predictions.}
\begin{tabularx}{\textwidth}{ZUUU}
\hline
\textbf{Model}                      & \textbf{Top-1} & \textbf{Top-5} & \textbf{Top-10} \\ \hline \hline
\multicolumn{4}{c}{\textbf{16 Frames}}                                                  \\ \hline
base-patch16-16-frames              & 0.06           & 0.28           & 0.39            \\
large-patch14-16-frames             & 0.05           & 0.22           & 0.42            \\
base-patch32-16-frames              & {\ul 0.08}     & {\ul 0.27}     & {\ul 0.55}      \\
base-patch16-kinetics-600-16-frames & 0.06           & 0.26  & 0.45            \\
\textbf{LoRA-16-frames-zero-shot}             & \textbf{0.12}  & \textbf{0.42}  & \textbf{0.62}   \\ \hline
\multicolumn{4}{c}{\textbf{8 Frames}}                                                   \\ \hline
base-patch16-kinetics-600           & 0.06           & 0.25           & 0.46            \\
large-patch14-kinetics-600          & 0.05           & 0.23           & {\ul 0.52}      \\
large-patch14                       & 0.06     & 0.24           & 0.46            \\
base-patch32                        & 0.05           & 0.26           & 0.51            \\
base-patch16                        & {\ul 0.07}     & {\ul 0.31}     & 0.46            \\
\textbf{LoRA-8-frames-zero-shot}              & \textbf{0.14}  & \textbf{0.40}  & \textbf{0.55}   \\ \hline
\end{tabularx}
\label{tab:zero_shot_metrics}
\end{table*}

\section{Conclusions and Future Work}
\label{sec:conclusions_and_future_work}

In this work, through an intense data cleaning process, we produced rich and aligned video-text pairs of capuchin monkey behavior from noisy raw video data, without relying on annotations. Part of these pairs were enriched with manual annotations and used to evaluate LoRA fine-tuned versions of X-CLIP for 16 and 8 frames against several different raw pre-trained configurations of the model. %With the obtained clean samples, we derived both training and an out-of-sample datasets, the latter enriched with manual annotations. We then evaluated LoRA fine-tuned versions of X-CLIP for 16 and 8 frames against several different raw pre-trained configurations of the model.

The choice of the models was based on their availability as open-source and also on their hardware requirements. For processing the textual transcripts, we used LLaMA 3.2 with 11B, while for video-text correlations we used BLIP-2 contrastive module, as an isolated data processing step. Hardware limitations not only affected the choice of models but also the fine-tuning strategies. For instance, layers to be fine tuned has been limited since fine-tuning initial (bottom) layers would require more memory.

The raw pre-trained models performed poorly: they were unable to rank most of the considered behaviors, as confirmed by metrics such as $NDCG@K$, and their retrieval results ($Hits@K$) were often close to random performance. In contrast, our method produced substantial gains in both ranking and retrieval, as proven by the computed metrics and supported by qualitative evaluation.

%Although we used LoRA as our fine-tuning method, which is a technique that usually performs well even with limited data, it is evident that our data processing method was so abrupt that we ended up with only a short number of samples. This is because, at this work, we only had access to a subsample of the data produced by LEDIS group. Even though the raw dataset was not small, it was also not large enough to produce a substantial volume of samples after cleaning, which imposed a limitation on the work.

Despite the relatively limited number of clip-transcript instances used in the fine-tuning process and the limitations in the model size and fine-tuning process, the experimental results suggest that the proposed methods are promising. In particular, we highlight that our method  does not require manual annotations and therefore it can be easily scaled to a large volume of videos or other videos with similar content.

%We believe that by using a larger multimodal LLM model for both text and video processing related steps in the data treatment pipeline has the potential to improve the results. This is even more true considering that we only used part of the available raw videos.

%Data size was not the only limitation that we faced in this work. Computational resources were also a significant limiting factor that influenced the model decisions we made. Because we are running open-source models locally, our hardware limited, for instance, the set of LLMs we could use. While we were able to adopt LLaMA 3.2 with 11B, which is not a small model, we only used the model to process the textual transcriptions, and not the video. Video-text correlations were later computed using BLIP-2 contrastive module, as an isolated data processing step. Probably, using both the text and the frames in the MLLM could have helped improve the quality of the treatment pipeline.

%The computational resources also affected our fine-tuning setup. If we choose too earlier layers to fine-tune, backpropagation go through more layers and, therefore, memory requirements are greater. Since most pre-trained models such as X-CLIP are huge, we were not able to fine-tune initial layers. In addition, few open-source pre-trained models are available in the context of video-text contrastive model, which limited the number of experiments we could explore.

Given that we were able to obtain reasonable results with the resources we had available, and in light of the highlighted limitations, some improvements can be pointed out for future work. First, in a near future we'll have access to a larger sample of the dataset, which will allow producing a larger volume of clean data. That will not only help us improve fine-tuning and allow the comparison of different fine-tuning methods, but also open the possibility for training models from scratch. Training our own models may also allow us to preserve the transcripts in the original language, without translating them to English. Finally, it is our intention to improve the data processing pipeline, extending the agent to contain reasoning capabilities and also including visual properties. Improving the agent may allow us to use smaller MLLMs, that require lower computational resources. On the application side, we intend to evaluate the usefulness of the developed methods to facilitate research.

\section*{Declaration of Competing Interest}
The authors declare that they have no known competing financial interests or personal relationships that could have appeared to influence the work reported in this paper.

\section*{Acknowledgment}
This research was partly supported by São Paulo Research Foundation (FAPESP), grants \#2022/15304-4, \#14/13237-1, \#21/11269-7, and \#2021/08153-7 and by MCTI-Brazil (law 8.248, PPI-Softex - TIC 13 - 01245.010222/2022-44).

\FloatBarrier

%% Appendix
\appendix
\section{Impact of LoRA Parameters}
\label{appendix:LoRA_params}

\setcounter{figure}{0}

\begin{figure*}[ht]
    \centering
    \includegraphics[width=0.3\textwidth]{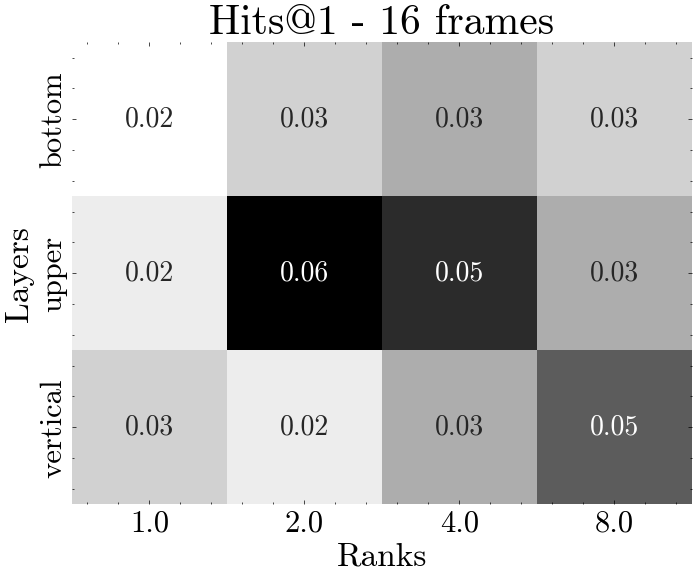}
    \hspace{0.01\textwidth}
    \includegraphics[width=0.3\textwidth]{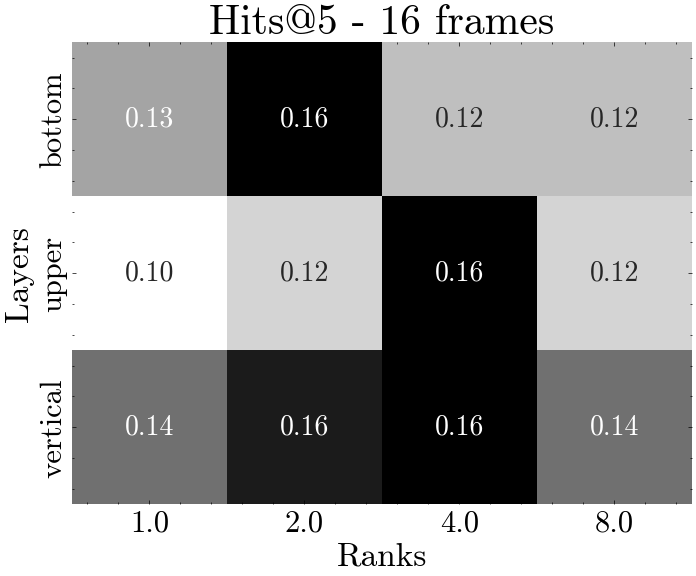}
    \hspace{0.01\textwidth}
    \includegraphics[width=0.3\textwidth]{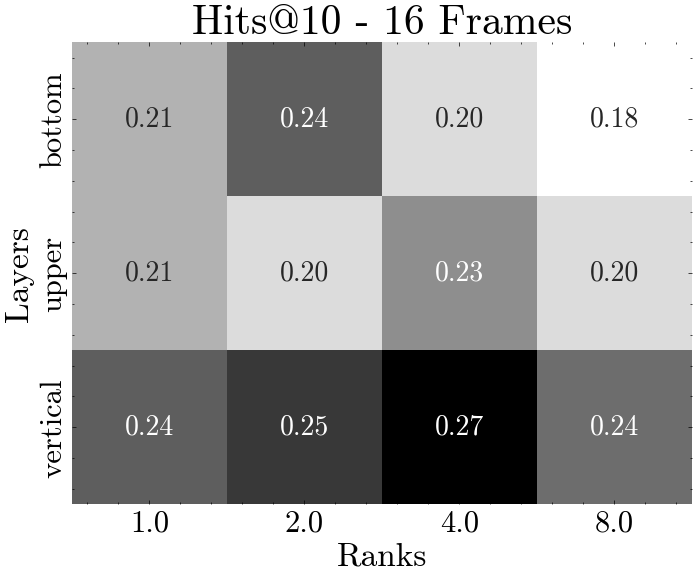}
    \\
    \vspace{0.01\textwidth}
    \includegraphics[width=0.3\textwidth]{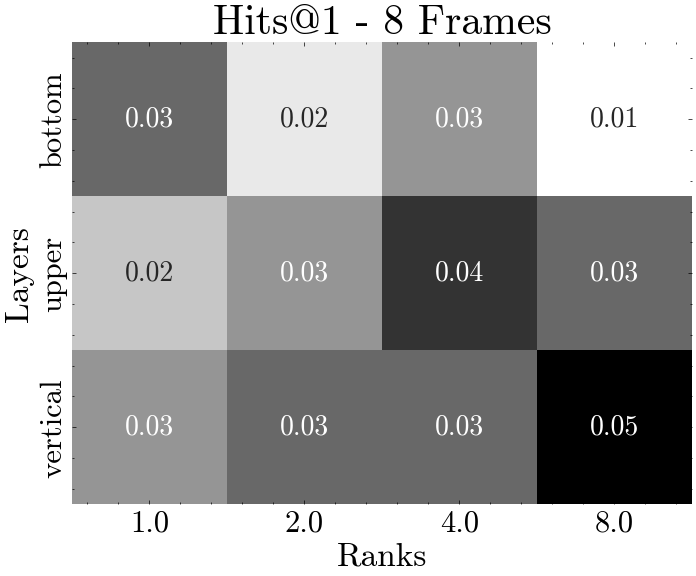}
    \hspace{0.01\textwidth}
    \includegraphics[width=0.3\textwidth]{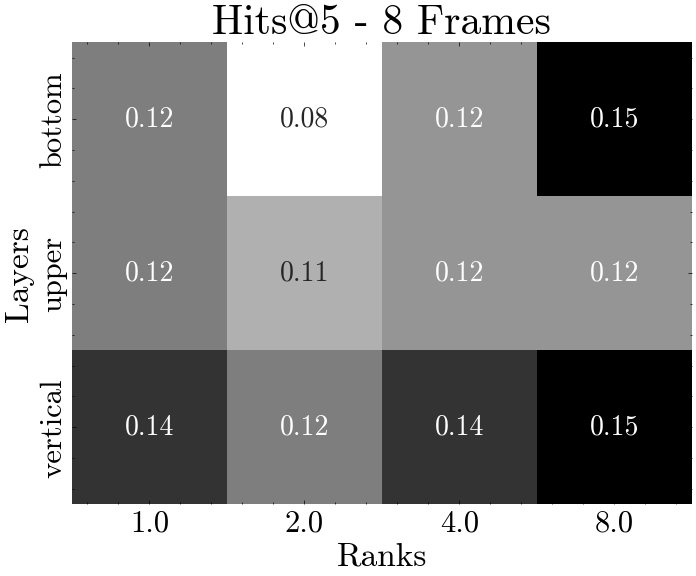}
    \hspace{0.01\textwidth}
    \includegraphics[width=0.3\textwidth]{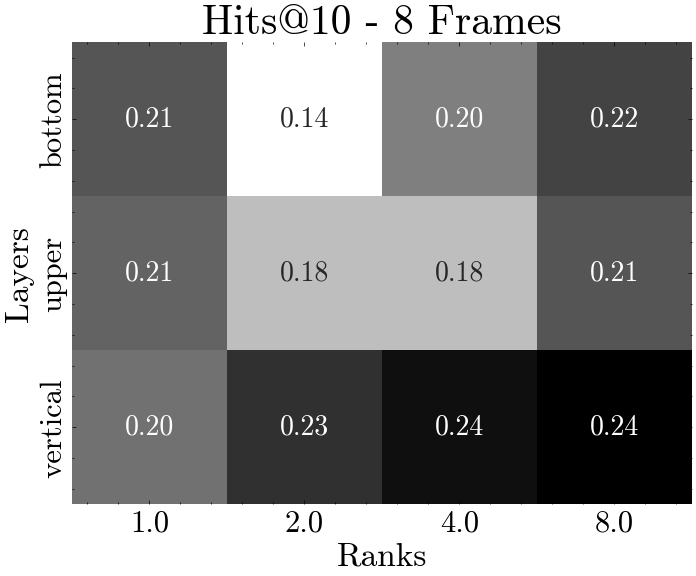}
    \caption{Effect of LoRA rank and model layers in the retrieval ($Hits@K$) metrics.}
    \label{fig:LoRA_matrix_retrieval}
\end{figure*}

\begin{figure*}[ht]
    \centering
    \includegraphics[width=0.3\textwidth]{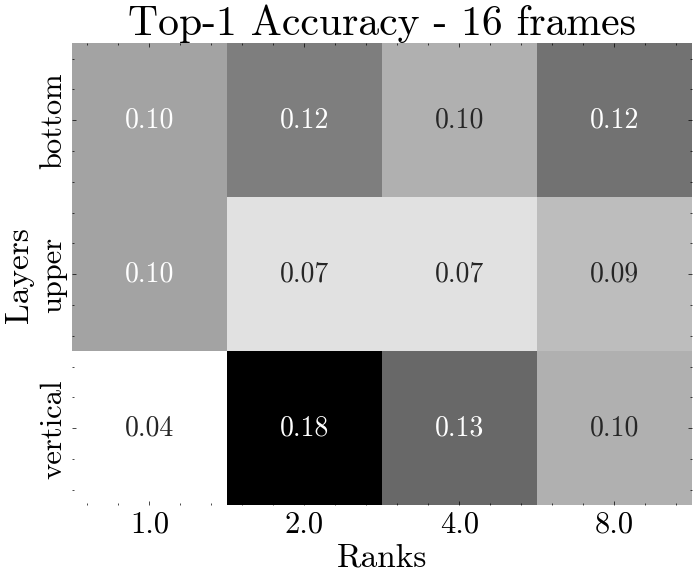}
    \hspace{0.01\textwidth}
    \includegraphics[width=0.3\textwidth]{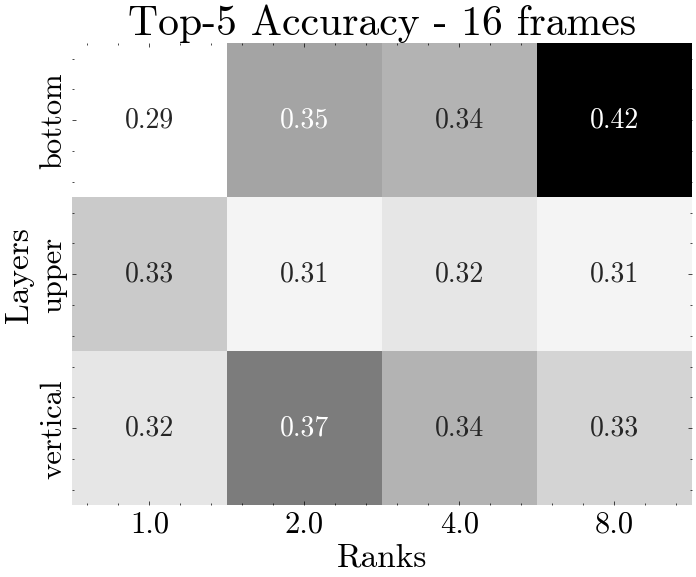}
    \hspace{0.01\textwidth}
    \includegraphics[width=0.3\textwidth]{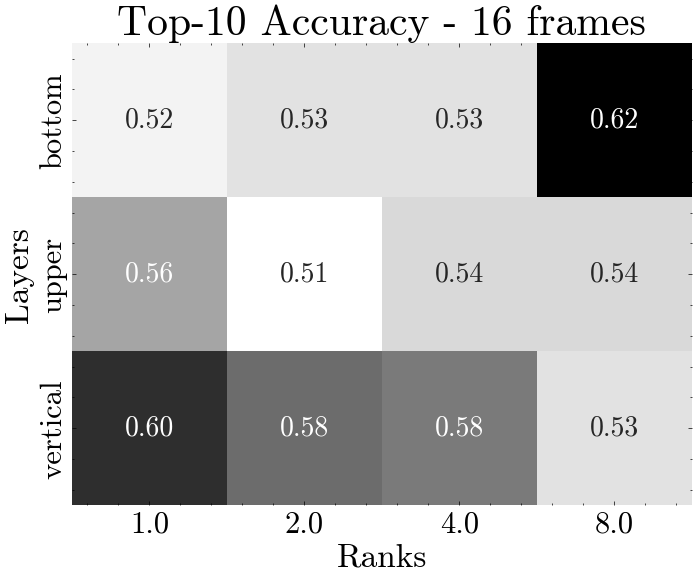}
    \\
    \includegraphics[width=0.3\textwidth]{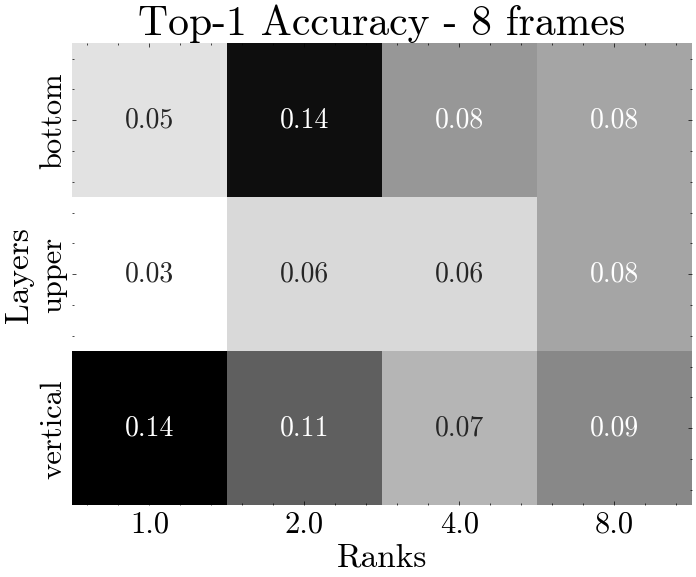}
    \hspace{0.01\textwidth}
    \includegraphics[width=0.3\textwidth]{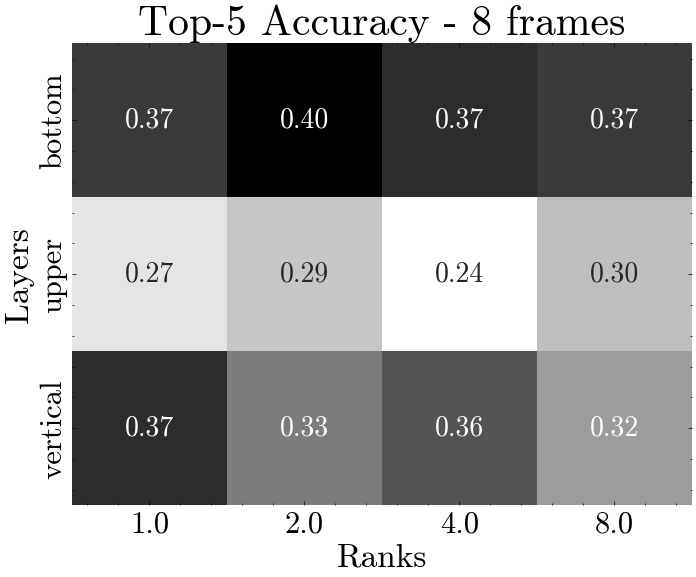}
    \hspace{0.01\textwidth}
    \includegraphics[width=0.3\textwidth]{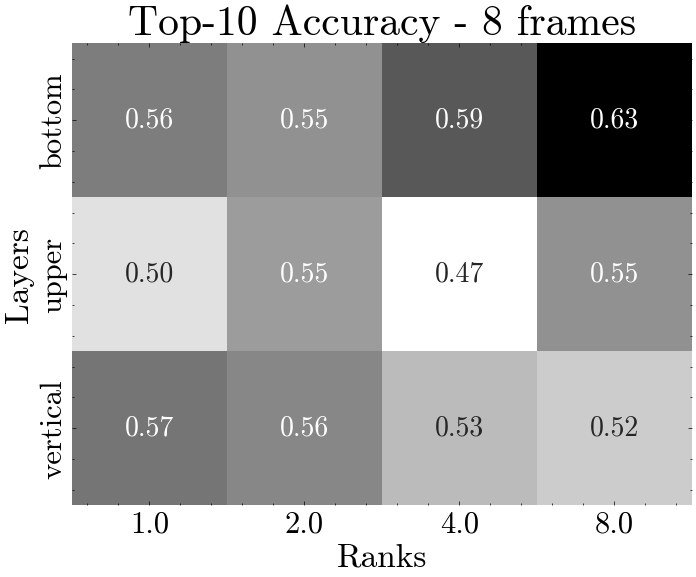}
    \caption{Effect of LoRA rank and model layers in Zero-shot accuracy.}
    \label{fig:LoRA_matrix_zero_shot}
\end{figure*}

\FloatBarrier

\bibliographystyle{elsarticle-num} 
\bibliography{bibdatabase}

\end{document}